\documentclass[acmsmall,natbib=false]{acmart}
\usepackage[style=ACM-Reference-Format,backend=bibtex,sorting=none]{biblatex}
\usepackage[T5]{fontenc}

\usepackage[utf8]{inputenc}

\usepackage[normalem]{ulem}
\usepackage{graphicx}
\usepackage{hyperref}
\usepackage{cleveref}
\usepackage{multirow}
\usepackage{multicol}
\usepackage{longtable}

\usepackage{pgfplots}

\usepackage{amsmath}

\usepackage{tgpagella}
\usepackage{biblatex}

\usepackage{verbatim} 

\usepackage{adjustbox}

\usepackage{tikz} 
\usetikzlibrary{arrows}

\usepackage{tabularx, makecell, booktabs}

\definecolor{bblue}{HTML}{4F81BD}

\usetikzlibrary{
    patterns,
}
    \pgfplotsset{
        compat=1.7,
        my ybar legend/.style={
            legend image code/.code={
                \draw [##1] (0cm,-0.6ex) rectangle +(2em,1.5ex);
            },
        },
}

\AtBeginDocument{%
  \providecommand\BibTeX{{%
    \normalfont B\kern-0.5em{\scshape i\kern-0.25em b}\kern-0.8em\TeX}}}

\setcopyright{acmcopyright}
\copyrightyear{2020}
\acmYear{2020}
\acmDOI{10.1145/1122445.1122456}

\begin{document}

\title{New Vietnamese Corpus for Machine Reading Comprehension of Health News Articles}


\author{Kiet Van Nguyen}
\affiliation{\institution{University of Information Technology, VNU-HCM, Vietnam}}
\author{Tin Van Huynh}
\affiliation{\institution{University of Information Technology, VNU-HCM, Vietnam}}
\author{Duc-Vu Nguyen}
\affiliation{\institution{University of Information Technology, VNU-HCM, Vietnam}}
\author{Anh Gia-Tuan Nguyen}
\affiliation{\institution{University of Information Technology, VNU-HCM, Vietnam}}
\author{Ngan Luu-Thuy Nguyen}
\affiliation{\institution{University of Information Technology, VNU-HCM, Vietnam}}



\renewcommand{\shortauthors}{Kiet Van Nguyen, et al.}
\renewcommand{\shorttitle}{New Vietnamese Corpus for Machine Reading Comprehension of Health News Articles}

\begin{abstract}
 Large-scale and high-quality corpora are necessary for evaluating machine reading comprehension models on a low-resource language like Vietnamese. Besides, machine reading comprehension (MRC) for the health domain offers great potential for practical applications; however, there is still very little MRC research in this domain. This paper presents ViNewsQA as a new corpus for the Vietnamese language to evaluate healthcare reading comprehension models. The corpus comprises 22,057 human-generated question-answer pairs. Crowd-workers create the questions and their answers based on a collection of over 4,416 online Vietnamese healthcare news articles, where the answers comprise spans extracted from the corresponding articles. In particular, we develop a process of creating a corpus for the Vietnamese machine reading comprehension. Comprehensive evaluations demonstrate that our corpus requires abilities beyond simple reasoning, such as word matching and demanding difficult reasoning based on single-or-multiple-sentence information. We conduct experiments using different types of machine reading comprehension methods to achieve the first baseline performances, compared with further models' performances. We also measure human performance on the corpus and compared it with several powerful neural network-based and transfer learning-based models. Our experiments show that the best machine model is ALBERT, which achieves an exact match score of 65.26\% and a F1-score of 84.89\% on our corpus. The significant differences between humans and the best-performance model (14.53\% of EM and 10.90\% of F1-score) on the test set of our corpus indicates that improvements in ViNewsQA could be explored in the future study. Our corpus is publicly available on our website\footnote[1]{https://sites.google.com/uit.edu.vn/uit-nlp/datasets-projects} for the research purpose to encourage the research community to make these improvements.
\end{abstract}


\maketitle

\section{Introduction}
\label{intro}

Question answering (QA) systems have recently achieved considerable success in a range of benchmark corpora due to the powerful development of neural network-based \cite{Che:2017,wang2018r, gupta2019deep} QA systems. Modern QA systems have two main components \cite{Che:2017}, where the first component for information retrieval selects text passages that appear relevant to questions from the corpus, and the second component for machine reading comprehension extracts answers that are then returned to the user. Machine Reading Comprehension (MRC) is a natural language understanding task that requires computers to understand human languages and answer questions by reading a given document. Human annotation for large-scale corpora is laborious and time-consuming, but more qualitative than data generation by automated method. Therefore, it is the best option for building many high-quality datasets such as SQuAD \cite{Raj:16} and CMRC \cite{Cui:18}. In order to evaluate MRC models, gold-standard resources comprising document-question-answer triples have to be collected and annotated by humans. Therefore, creating a benchmark corpus is vital for human language processing, especially for low-resource languages such as Vietnamese.

In recent years, researchers have developed many MRC corpora and models in popular languages such as English and Chinese. The best-known examples of gold standard MRC resources for English are span-extraction MRC corpora \cite{Raj:16,Raj:18,Tri:16}, cloze-style MRC corpora \cite{Her:15,Hil:15,Cui:16}, reading comprehension with multiple-choice \cite{Ric:13,Lai:17}, and conversation-based reading comprehension \cite{Red:19,Sun:19}. Examples of the resources available for other languages include the Chinese corpus for the span-extraction MRC \cite{Cui:18}, traditional Chinese corpus of MRC \cite{Shao:18}, the user-query-log-based corpus DuReader \cite{He:17}, and the Korean MRC corpus \cite{Lim:19}. In addition to development of the reading comprehension corpora, various significant neural network-based approaches have been proposed and made a significant advancement in this research field, such as Match-LSTM \cite{Wan:16}, BiDAF \cite{Seo:16}, R-Net \cite{Wang:17}, DrQA \cite{Che:2017}, FusionNet \cite{Hua:17}, FastQA \cite{Wei:17}, QANet \cite{Yu:18}, and S3-NET \cite{par:20}. Powerful transfer learning models such as BERT \cite{Dev:18} and its variants (ALBERT \cite{lan2019albert}) have recently become extremely popular and achieved state-of-the-art results in MRC tasks.


Although researchers have studied several works on the Vietnamese language, such as parsing \cite{nguyen2009building,nguyen2014treebank,nguyen2016x,nguyen2018lstm}, part-of-speech \cite{datpos,bachpos}, named entity recognition \cite{10.1145/1316457.1316460,10.1145/2990191,nguyen2019error}, sentiment analysis \cite{van2018uit,nguyen2018deep,van2018transformation}, and question answering \cite{Nguyen_2009,van2016improving,le2018factoid}, there is only two corpora for evaluating MRC models, ViMMRC \cite{Ngu:20} for evaluating Vietnamese multiple-choice questions and UIT-ViQuAD \cite{Kiet:20ViQuAD} for evaluating Vietnamese span-extraction MRC models. However, both two corpora are open-domain. In this paper, we aim to build a new large Vietnamese corpus based on online news articles in the health domain for evaluating MRC models. There are several main reasons for this. Firstly, machine comprehension for health domain has few studies so far, although it could be implemented into various potential and practical applications such as chatbot and virtual assistant in health-care service. Secondly, this study aims to build an application for general readers who search information and health-domain knowledge from online health articles. Finally, a new corpus is our important contribution to assess different MRC and QA models in a low-resource Vietnamese language.



The current approaches based on deep neural networks and transfer learning have surpassed the performance of humans with English corpora like SQuAD, but it is not clear these state-of-the-art models will obtain similar performance with corpora in different languages. Hence, to further enhance the development of the MRC, we develop a new span-extraction corpus for Vietnamese MRC. In this paper, we have three main contributions described as follows.
\begin{itemize}
    \item Firstly, we develop a benchmark corpus (ViNewsQA) for evaluating Vietnamese machine reading comprehension and question answering systems. ViNewsQA comprises over 22,000 human-created question-answer pairs based on over 4,400 online news articles in the health domain. The corpus is publicly available for Vietnamese language processing research and also for the cross-lingual studies together with other similar corpora such as NewsQA (for English), CMRC (for Chinese), FQuAD (for French) and KorQuAD (for Korean).
    \item Besides, we analyze the corpus in terms of different linguistic aspects, including  vocabulary-based, three types of length (question, answer, and article), three content-based types (question, answer and reasoning) and the correlation between type-based and the answer length, thereby providing comprehensive insights into the corpus that may facilitate future methods.

    \item Finally, we conduct the first experiments on different types of MRC methods as the first baseline models on the ViNewsQA corpus. The best-performance baseline is ALBERT with 65.26\% (in EM) and 84.89\% (in F1-score). The significant difference between humans and the best-performance model (10.90\% of F1-score) indicates that improvements in ViNewsQA could be explored in the future study. In addition, we compare their performances with humans in terms of various linguistic aspects to obtain in-depth insights into Vietnamese span-extraction machine reading comprehension in the health domain using different methods.
\end{itemize}

The remainder of this paper is structured as follows. In Section 2, we review the existing machine reading comprehension corpora and models. In Section 3, we explain the creation process of our corpus. The analysis of our corpus is described in Section 4. Then, we present our experimental evaluation (in Section 5) and analysis of the experimental results and discussion (in Section 6). Finally, we draw our conclusions and suggest directions for future research in Section 7.

\section{Related Work}
\label{relate}
In this section, we review several studies related to our work, including related MRC corpora (in Section 2.1) and models (in Section 2.2).
\subsection{Related MRC Corpora}
 Depending on the answer, the answers are divided into different types. Generally, reading comprehension tasks can be divided into four classes: cloze style \cite{xie-etal-2018-large,cui-etal-2020-sentence}, multiple choice \cite{Ric:13,Lai:17}, span extraction \cite{Raj:16,Cui:18} and free form \cite{yang2019end}. In this study, we aim to construct a span-extraction MRC corpus on the health-domain online news for the Vietnamese language. We review thoroughly the corpora related to the span-extraction MRC and the health domain. Our ViNewsQA corpus is inspired by various recent span-extraction reading comprehension corpora, such as SQuAD \cite{Raj:16}, NewsQA \cite{Tri:16}, CMRC \cite{Cui:18}, and KorQuAD \cite{Lim:19}. In particular,  CliCR \cite{vsuster2018clicr}, MedQA \cite{zhang2018medical} and PubMedQA \cite{pubmedqa} are three first MRC corpora in the health domain created in 2018. Table \ref{tab:datasetreview} summarize a brief of these corpora.

\begin{table}[H]
\caption{A survey of several corpora related to our corpus ViNewsQA.}
\label{tab:datasetreview}
\setlength\arrayrulewidth{1pt}
\resizebox{\textwidth}{!}{\begin{tabular}{lcccrc}
\hline
\textbf{Corpus}                                                        & \textbf{Language}   & \textbf{Domain}         & \textbf{Type}            & \multicolumn{1}{c}{\textbf{Size}} & \textbf{Annotation Method} \\ \hline
SQuAD \cite{Raj:16}                                                                   & English             & Open                    & \textbf{Span-extraction} & 100K                               & Crowdsourcing              \\ 
NewsQA \cite{Tri:16}                                                                  & English             & Open                    & \textbf{Span-extraction} & 100K                               & Crowdsourcing              \\ 
CMRC \cite{Cui:18}                                                                    & Chinese             & Open                    & \textbf{Span-extraction} & 20K                                & Crowdsourcing              \\ 
KorQuAD \cite{Lim:19}                                                                 & Korean              & Open                    & \textbf{Span-extraction} & 70K                                & Crowdsourcing              \\ 
FQuAD \cite{Mar:20}                                                                   & French              & Open                    & \textbf{Span-extraction} & 60K                                & Crowdsourcing              \\ 
SberQuAD \cite{Pav:20}                                                               & Russian             & Open                    & \textbf{Span-extraction} & 90K                                & Crowdsourcing              \\ 
UIT-ViQuAD \cite{Kiet:20ViQuAD}                                                             & \textbf{Vietnamese} & Open                    & \textbf{Span-extraction} & 23K                                & Crowdsourcing              \\ 
ViMMRC \cite{Ngu:20}                                                                 & \textbf{Vietnamese} & Open                    & Multiple-choice          & 2.7K                               & Crowdsourcing              \\ 
CliCR \cite{vsuster2018clicr}                                                                  & English             & \textbf{Medical}        & Gap-filling              & 105K                               & Crowdsourcing            \\ 
MedQA \cite{zhang2018medical}                                                                   & English             & \textbf{Medical}        & Multiple-choice                      & 270K                                & Published materials                         \\ 
PubMedQA \cite{pubmedqa}                                                               & English             & \textbf{Medical}        & Yes/No                      & 273K                                & Crowdsourcing                        \\ \hline
\textit{\begin{tabular}[c]{@{}c@{}}ViNewsQA\\ (this work)\end{tabular}} & \textit{Vietnamese} & \textit{Medical + News} & \textit{Span-extraction} & \textit{22K}                       & \textit{Crowdsourcing}     \\ \hline
\end{tabular}}
\end{table}

Table \ref{tab:datasetreview} presents several MRC corpora and their characteristics. For the extraction-span MRC corpora, we review and analyze several well-known corpora, including SQuAD, NewsQA, CMRC, KorQuAD, FQuAD and SberQuAD. \textbf{SQuAD} is one of the best-known English corpora for the extractive MRC and it has facilitated the development of many machine learning models. In 2016, Rajpurkar et al. ~\shortcite{Raj:16} proposed SQuAD v1.1 comprising 536 Wikipedia articles with 107,785 human-generated question and answer pairs. SQuAD v2.0 \cite{Raj:18} was based on SQuAD v1.1 but it includes over 50,000 unanswerable questions created adversarially 
using the crowd-worker method according to the original questions. \textbf{NewsQA} is another English corpus proposed by Trischler et al. \shortcite{Tri:16}, which comprises 119,633 question-answer pairs generated by crowd-workers based on 12,744 articles from the CNN news. This corpus is similar to SQuAD because the answer to each question is a text segment of arbitrary length in the corresponding news article.  \textbf{CMRC} \cite{Cui:18} is a span-extraction corpus for Chinese MRC, which was introduced in the Second Evaluation Workshop on Chinese Machine Reading Comprehension in 2018. This corpus contains approximately 20,000 human-annotated questions on Wikipedia articles. This competition attracted many participants to conduct numerous experiments on this corpus. \textbf{KorQuAD} \cite{Lim:19} is a Korean corpus for span-based MRC, comprising over 70,000 human-generated question-answer pairs based on Korean Wikipedia articles. The data collected and the properties of the data are similar to those in the English standard corpus SQuAD. \textbf{FQuAD} \cite{Mar:20} is a French native reading comprehension corpus of questions and answers on a set of Wikipedia articles that consists of 25K questions for the 1.0 version and 60 questions for the 1.1 version. \textbf{SberQuAD} \cite{Pav:20} contains 50K paragraph–question–answer triples and was created in a similar way to SQuAD. SberQuAD selected Wikipedia pages, split into paragraphs, and paragraphs presented to crowd workers. For each paragraph, a Russian native speaking crowd worker posed questions that can be answered using solely the content of the paragraph and their answers must have been a paragraph span, i.e., a contiguous sequence of paragraph words. All of these corpora are built based on the crowdsourcing method, which has motivated to build our corpus.

For the Vietnamese language, there are only two corpora for evaluating MRC models, including ViMMRC \cite{Ngu:20} and UIT-ViQuAD \cite{Kiet:20ViQuAD}. \textbf{ViMMRC} is the first Vietnamese corpus which consists of 2,783 pairs of multiple-choice question-answer-passage triples which are commonly used for teaching reading comprehension for elementary school students. In addition, \textbf{UIT-ViQuAD} is a span-extraction open-domain corpus for the low-resource language as Vietnamese to evaluate MRC models. This corpus consists of over 23K human-generated question-answer pairs based on 5,109 passages of 174 Vietnamese articles from Wikipedia. Both of these corpora are open-domain, we want to target a  domain-specific and be useful for future practical applications.

We choose the health domain for our corpus. Hence, we review several related corpora in this domain. \textbf{CliCR} \cite{vsuster2018clicr} is a medical-domain corpus comprising around 100,000 gap-filling queries based on clinical case reports, while \textbf{MedQA} \cite{zhang2018medical} collected answer real-world multiple-choice questions with large-scale reading comprehension. These corpora required world and background domain knowledge in the study of the MRC models. {\bf PubMedQA} \cite{pubmedqa} is a novel biomedical QA corpus collected from PubMed abstracts with 273 yes/no/maybe QA instances. These corpora are mainly aimed at simple forms of English reading comprehension like filling-gap, multiple-choice and yes/no questions.



Until now, there are not any Vietnamese corpus for the span-based MRC research in the health-domain online news. The benchmark corpora mentioned above used for evaluating the MRC models and developing different QA applications, thereby encouraging researchers to explore the machine-learning models on these corpora. Our corpus is also intended for these purposes. These reasons lead to create a Vietnamese corpus in the health domain for MRC tasks.

\subsection{Related MRC Methods}

To the best of our knowledge, a range of studies have investigated MRC methodologies, and three popular approaches for MRC are rule-based, neural network-based and transfer learning-based. Rule-based approach is the first baseline of many well-known corpora \cite{Raj:16,Ric:13,Sun:19,Ngu:20}. However, neural network-based and transfer learning-based systems have recently become more prevalent in MRC systems due to the powerful development of large-scale and high-quality corpora. In particular, we review them in detail as follows.

{\bf Rule-based Approaches.}
Sliding window (SW) is the first rule-based approach developed by Richardson et al. (2013) \cite{Ric:13}. This approach matches a set of words built from a question and one of its answer candidates with a given reading text, before calculating the matching score using TF-IDF for each answer candidate. Experiments have been conducted with this simple model on many different corpora as first baseline models, such as MCTest \cite{Ric:13}, SQuAD \cite{Raj:16}, DREAM \cite{Sun:19}, and ViMMRC \cite{Ngu:20}.

{\bf Machine Learning-Based Approaches.}
 In addition to the rule-based models, machine-learning-based models have interesting features due to the development of large and high-quality corpora and robust machine configurations. In particular, Rajpurkar et al. \cite{Raj:16} introduced a logistic regression model with a range of different linguistic features. However, neural network-based models on this problem have attracted more attention and obtained outstanding results in recent years. The corpora mentioned in Sub-section 2.1 have been studied in the development and evaluation of various neural network-based models in the field of natural language processing, such as Match-LSTM \cite{Wan:16}, BiDAF \cite{Seo:16}, CNN-LR \cite{jurczyk2016selqa}, R-Net \cite{Wang:17}, DrQA \cite{Che:2017}, FusionNet \cite{Hua:17}, FastQA \cite{Wei:17}, QANet \cite{Yu:18}, and S3-NET \cite{par:20}. In recent years, transfer-learning models have shown their strengths on many NLP tasks. In perticular, Devlin et al. \shortcite{Dev:18}, Lan et al. \shortcite{lan2019albert},  and Conneau et al. \shortcite{Con:19} introduced BERT and its variants (ALBERT and XLM-R), respectively, as powerful models trained on multiple languages and they obtained state-of-the-art performance with machine reading comprehension corpora.
 

In this paper, we choose several typical methods from three popular types of MRC models comprising rule-based (Sliding Window), neural network-based (DrQA and QANet) and transfer learning-based (BERT and ALBERT) for our machine reading comprehension corpus. In addition, we attempt to analyze the experimental results in terms of different linguistic aspects to gain first insights into Vietnamese machine reading comprehension in the health domain.

\section{Corpus}
\label{data}
In this section, we introduce the task of machine reading comprehension and give several examples in Vietnamese (in Section \ref{definition}). Then, we present how to create a new corpus for evaluating Vietnamese machine reading comprehension in the health domain (in Section \ref{corpuscreation}). These sections are described as follows.

\subsection{Task Definition}
\label{definition}
Formally, the reading comprehension task is described as a triple $(D, Q, A)$, where $D$ represents a document, $Q$ represents a question, and $A$ means an answer. Documents in our corpus are online news articles. Specifically, for the span-based reading comprehension task, question-answer pairs are created by humans. The answer $A$ is a continuous span that is directly extracted from the document $D$. Figure \ref{tab:example} presents several examples for Vietnamese span-extraction reading comprehension in the health-domain online news.

\begin{table}[H]
\resizebox{\columnwidth}{!}{\begin{tabular}{p{15cm}}
\hline
\begin{tabular}[c]{p{15cm}}{\bf Document}: Nghiên cứu cho thấy resveratrol trong rượu vang đỏ có khả năng làm giảm huyết áp, khi thí nghiệm trên chuột. Resveratrol là một hợp chất trong vỏ nho {\color{red}{\bf có khả năng chống oxy hóa, chống nấm mốc và ký sinh trùng}}. Trên Circulation, các nhà khoa học từ King's College London (Anh) công bố kết quả thí nghiệm tìm ra sự liên quan giữa chuột và resveratrol. Cụ thể, {\color{blue}{\bf resveratrol tác động đến huyết áp của những con chuột này, làm giảm huyết áp của chúng}}.\\ ({\it The study showed that resveratrol in red wine could reduce blood pressure when tested in mice. Resveratrol is a compound found in grape skin that {\color{red} has antioxidant, anti-mold, and anti-parasitic properties}. Scientists from King’s College London (UK) published experimental results in Circulation regarding a link between mice and resveratrol. Specifically, {\color{blue} resveratrol affected the blood pressure of these mice, lowering their blood pressure}.})\end{tabular}
\\ \hline

\begin{tabular}[c]{p{15cm}}{\bf Question 1}: Chất bổ trong vỏ nho có tác dụng gì? ({\it What is the substance in grape skin for?})
\\{\bf Answer}: {\color{red}{\bf có khả năng chống oxy hóa, chống nấm mốc và ký sinh trùng}} ({\color{red}{\it has antioxidant, anti-mold, and anti-parasitic properties}}). \end{tabular}                                                                                                                    \\ \hline
\begin{tabular}[c]{p{15cm}}{\bf Question 2}: Các nhà khoa học từ trường King's tìm ra phát hiện gì về loài chuột và resveratrol? \\({\it What did scientists from King's University discover about mice and resveratrol?})\\{\bf Answer}: {\color{blue}{\bf resveratrol tác động đến huyết áp của những con chuột này, làm giảm huyết áp của chúng}} ({\color{blue}{\it resveratrol affected the blood pressure of these mice, lowering their blood pressure}}).\end{tabular}                                                                    \\ \hline
\end{tabular}}
\captionof{figure}{Several examples of our proposed corpus (ViNewsQA). English translations are also provided for comparison.} \label{tab:example}
\end{table}

\subsection{Corpus Creation}
\label{corpuscreation}

\begin{figure}[H]
\includegraphics[width=14cm]{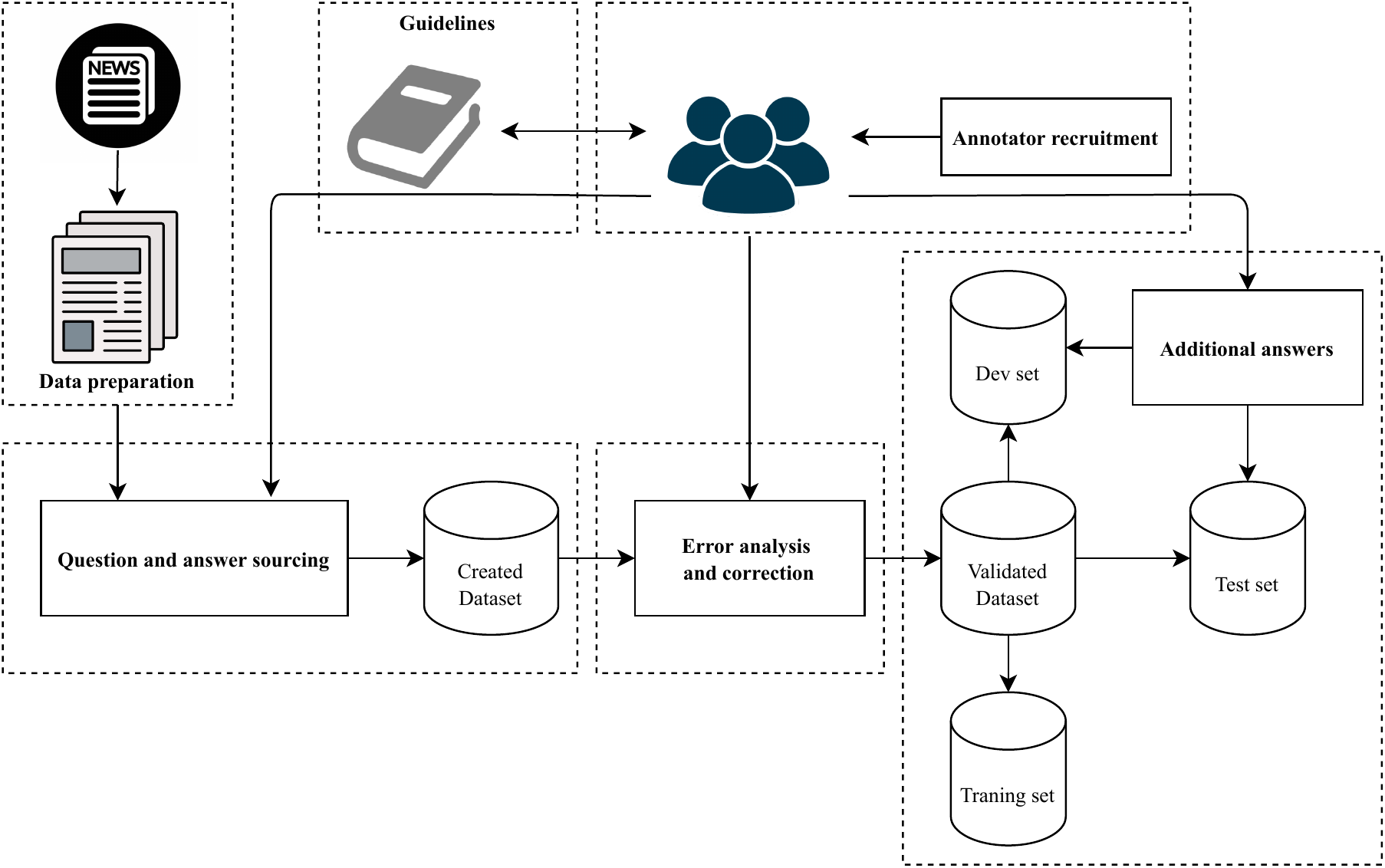}
\caption{The overview process of creating the Vietnamese MRC corpus in the heath domain.}\label{fig:Datasetcreation}
\centering
\end{figure}

In this section, we present a new process to create the Vietnamese MRC corpus in the health domain, as shown in Figure \ref{fig:Datasetcreation}. In particular, we construct our corpus through six different phases comprising (see in Section 3.2.1) annotator recruitment, (see in Section 3.2.2) building guidelines, (see in Section 3.2.3) data preparation, (see in Section 3.2.4) question and answer sourcing, (see in Section 3.2.5) validation based on error analysis and correction, and (see in Section 3.2.6) collecting additional answers. We describe these phases in detail as follows.

\subsubsection{Annotator recruitment}
We hire annotators to build our corpus according to a rigorous process in the following three different stages described as follows.

\begin{itemize}
    \item {\bf Stage 1}: People, who have an interest in reading health-domain online news, apply to become annotators to create the question-answer pairs for the MRC task. 
    \item {\bf Stage 2}: Annotators selected are good at general knowledge and passed our reading comprehension test.
    \item {\bf Stage 3}: Official annotators are carefully trained guidelines (see in Section \ref{guideline}) with 200 questions. They MUST follow annotation rules presented in Section \ref{guideline}.
\end{itemize}
\subsubsection{Guidelines}
\label{guideline}

The annotators read and understand each article, and they then formulate questions and select their answers directly in the article. During the creation process of question-answer pairs, the annotators conform to the following rules. 
\begin{itemize} 
\item {\bf Rule 1}: Annotators are required to pose at least three question-answer pairs per the article. 
\item {\bf Rule 2}: Annotators are encouraged to ask questions in their own words and vocabulary. 
\item {\bf Rule 3}: The answer MUST be a span in the article that satisfy the requirements of the task definition. The spans with the shortest length from potential answers are encouraged to be selected for the answers to the questions. 
\item {\bf Rule 4}: To diverse different types of questions, annotators are encouraged to create questions with different types (what/who/when/where/why/how, etc.). In addition, complex reasoning (single-sentence and multiple-sentence reasoning) is also encouraged in the question generation.
\item {\bf Rule 5}: Annotators are warned about mistakes that could be avoided when creating questions–answer pairs. These mistakes are shown from our error analysis presented in Section \ref{validation}.
\end{itemize}

\subsubsection{Data preparation}
4,416 news articles related to {\bf the health topic  are collected from the online newspaper} VnExpress\footnote[4]{https://vnexpress.net/suc-khoe}. {\bf We choose this source because it is one of the most popular Vietnamese online newspaper\footnote[5]{https://en.wikipedia.org/wiki/VnExpress}\footnote[6]{https://www.alexa.com/topsites/countries/VN} and the language used in articles is easy to understand for general readers, which aim for practical applications.} All images, figures, and tables are eliminated from these articles, and articles shorter than 300 characters or those containing many special characters and symbols are removed. We divide the articles randomly into a training set (Train), a development set (Dev), and a test set (Test) with an approximate rate of 8:1:1 {\bf for conducting experiments on machine reading comprehension models in Section \ref{experiment}}.
\subsubsection{Question and answer sourcing}
\label{annotationrule}
Following the guidelines (see in Section \ref{guideline}), annotators create question-answer pairs per article. Annotators use the MRC annotation tool that we build to create question-answer pairs. In each working section, the tool allows to display the article content and enables the annotators to enter questions and choose their answers directly on the article and also allows the annotators to save the article content, the questions, and answers on a *.json file.

\subsubsection{Error analysis and correction}
\label{validation}

Errors that may arise when manually creating questions and choosing answers from articles are inevitable. To enhance the quality of the corpus, we perform the validation process to minimize these errors. To analyze the error types that can occur during the data generation of annotators, we select randomly 1,183 question-answer pairs (over 5\% of the corpus) to investigate the errors and find 335 question-answer pairs with mistakes. Based on questions or answers, we divide these errors into five different types such as unclear questions (Error type 1), misspelled questions (Error type 2), incorrect answers (Error type 3), lack-or-excess-of-information answers (Error type 4), and incorrect-boundary answers (Error type 5). These errors are described as follows.

\begin{itemize}
    \item {\bf Error type 1}: Questions are misspelled. In the process of creating questions and their answers, annotators could misspell during the typing process. 
    \item {\bf Error type 2}: Answers are incorrect for their questions. In particular, questions are correct, but their selected answers are wrong.
    \item {\bf Error type 3}: Answers are lack or excess of information for questions. In particular, annotators can choose the redundant or unnecessary text to answer their questions.
    \item {\bf Error type 4}: Questions are not precise and clear in their contents. People cannot understand these questions, so they do not find answers to these questions.
    \item {\bf Error type 5}: Answers are incorrect-boundary spans. Remarkably, the annotators can choose either lack or excess some characters or spaces in the answer.
    
\end{itemize}

Table \ref{tab:errortype} presents the common types of errors that annotators made during the corpus creation process. We find that the error type 3 occurs most frequently and accounts for 54.33\% while the error type 5 accounts for the lowest percentage of 1.49\%. From these analyses, we require the annotators to check and correct carefully with these errors. Besides, these types of errors are useful for future development of MRC corpora.

\begin{longtable}[c]{cp{9cm}p{1.8cm}}
\caption{Statistics of error types of annotators when creating question-answer sourcing. Examples and their English translations are also given for comparison.}
\label{tab:errortype}
\\
\hline
\multicolumn{1}{c}{\textbf{Error types}} &
  \textbf{Examples} &
  \multicolumn{1}{c}{\textbf{Percentage (\%)}} \\ \hline
\endfirsthead
\endhead
1 &

  \begin{tabular}[c]{p{8.5cm}}\textbf{Annotator’s question:} Salmonella gây ra những \underline{gnuy hiểm} gì cho phụ nữ mang thai?\\ \textbf{Correct question:} Salmonella gây ra những \underline{nguy hiểm} gì cho phụ nữ mang thai?\\\\ \textbf{English translation:}\\ \textbf{Annotator’s question:} What are the \underline{adngers} of Salmonella in pregnant women?\\ \textbf{Correct question:} What are the \underline{dangers} of Salmonella in pregnant women?\end{tabular} 
  & \multicolumn{1}{r}{4.78}\\ \hline
2 &

  \begin{tabular}[c]{p{8.5cm}}\textbf{Document:} Bệnh thủy đậu xảy ra ở mọi lứa tuổi, chủ yếu ở trẻ em. Những người có hệ miễn dịch kém như người trên 50 tuổi, suy dinh dưỡng hoặc đang sử dụng thuốc điều trị ung thư, thuốc ức chế miễn dịch, phụ nữ có thai... có nguy cơ cao mắc bệnh. Người lớn thường bị trong các trường hợp suy giảm miễn dịch, thông thường bệnh sẽ nặng hơn ở trẻ em.\\ \textbf{Annotator’s question:} Nếu mắc bệnh thuỷ đậu thì ai sẽ mắc bệnh nặng hơn?\\ \textbf{Annotator’s answer:} Trẻ em.\\ \textbf{Correct answer:} Người lớn.\\\\ \textbf{English translation:}\\ \textbf{Document:} Chickenpox happens to people of all ages, mostly in children. People with weakened immune systems such as those aged over 50, malnourished or taking cancer treatment drugs, immunosuppressants, pregnant women, etc. have a greater incident of the disease. Adults are often infected in cases of immunodeficiency while the children suffer more seriously.\\ \textbf{Annotator’s question:} Assuming someone gets the chickenprox, who could be worse?\\ \textbf{Annotator’s answer:} Children.\\ \textbf{Correct answer:} Adult.\end{tabular} & \multicolumn{1}{r}{4.78}\\ \hline
3 &

  \begin{tabular}[c]{p{8.5cm}}\textbf{Document:} Hydro sulfua là hợp chất khí ở điều kiện nhiệt độ thường, không màu, mùi trứng thối.\\ \textbf{Annotator’s question:} Tính chất vật lý của khí H2S là gì?\\ \textbf{Annotator’s answer:} hợp chất khí ở điều kiện nhiệt độ thường.\\ \textbf{Correct answer:} hợp chất khí ở điều kiện nhiệt độ thường, không màu, mùi trứng thối.\\\\ \textbf{English translation:}\\ \textbf{Document:} Hydrogen sulfide is a gas compound at normal temperature conditions, colorless, rotten egg smell.\\ \textbf{Annotator’s question:} What are the physical properties of H2S?\\ \textbf{Annotator’s answer:} gas compound at room temperature.\\ \textbf{Correct answer:} gas compound at normal temperature, colorless, rotten egg smell.\end{tabular} & \multicolumn{1}{r}{54.33}\\ \hline
  
4 &
  
  \begin{tabular}[c]{p{8.5cm}}\textbf{Document:} Mọi người cần tiêm phòng thuỷ đậu cho tất cả trẻ em và người lớn chưa nhiễm bệnh. Hiện vắcxin này được sử dụng khá phổ biến và đem lại hiệu quả cao trong việc phòng bệnh, ít gây tác dụng phụ.\\ \textbf{Annotator’s question:} Vắcxin hiện nay có được sự cải tiến nào?\\ \textbf{Correct question:} Vắcxin thủy đậu hiện nay có được sự cải tiến nào?\\\\ \textbf{English translation:}\\ \textbf{Document:} People are in need of vaccinating all uninfected children and adults against the chickenpox. Currently, this vaccine is used quite commonly and brings about the high efficiency in the prevention of disease while causing few side effects.\\ \textbf{Annotator’s question:} Is there any improvement in current vaccines?\\ \textbf{Correct question:} Which improvement is being taken place on the current chickenpox vaccines?\end{tabular} & \multicolumn{1}{r}{34.62}\\ \hline
5 &

  \begin{tabular}[c]{p{8.5cm}}\textbf{Document:} Trong 101.862 trẻ được tiêm vắcxin ComBE Five trên 19 tỉnh, ghi nhận 1,73\% trẻ có phản ứng thông thường như sốt nhẹ, sưng đau tại chỗ tiêm, khó chịu, quấy khóc.\\ \textbf{Annotator’s answer:} ốt nhẹ, sưng đau tại chỗ tiêm, khó chịu, quấy khóc.\\ \textbf{Correct answer:} sốt nhẹ, sưng đau tại chỗ tiêm, khó chịu, quấy khóc.\\\\ \textbf{English translation:}\\ \textbf{Document:} Among 101,862 children were injected the ComBE Five vaccine in 19 provinces, 1.73\% of them have some slightly reaction such as low fever, soreness at the injection site, discomfort, and crying.\\ \textbf{Annotator’s answer:} ow fever, soreness at the injection site, discomfort, crying.\\ \textbf{Correct answer:} low fever, soreness at the injection site, discomfort, crying.\end{tabular} & \multicolumn{1}{r}{1.49}\\ \hline
\end{longtable}

\subsubsection{Collecting additional answers}
\label{additionalanswer}
To estimate the performance of humans (see in Section \ref{humper}) and to enhance our empirical evaluations of the development and test sets, we add two more answers for each question, and the first answers in the development and test sets are annotated by annotators. During this phase, the annotators do not know the first answer, and they are encouraged to give diverse answers.


\section{Corpus Analysis}
\label{datasetanalysis}

Firstly, we introduce the overview of our corpus in Section \ref{overviewsta}. To understand the characteristics of our corpus ViNewsQA, we perform a variety of analyzes based on linguistic aspects such as vocabulary-based in Section \ref{vocab}, length-based (question length, answer length and article length) in Section \ref{lengthana} and type-based (question type, answer type and reasoning type) in Section \ref{typeana}. In addition, we analyze the correlation between type-based and the answer length in Section \ref{correlation}. These analyses provide in-depth insights into our corpus and comparisons with another Vietnamese corpus as UIT-ViQuAD. For question type, answer type and reasoning type, we also hire annotators to annotate questions of the development set which are selected randomly from our corpus. Corpora such as SQuAD \cite{Raj:16} and UIT-ViQuAD \cite{Kiet:20ViQuAD} also perform corpus analysis on the development set.


\subsection{Overall statistics}
\label{overviewsta}
Before conducting detailed analyses, we provide an overview of our corpus. Table \ref{tab:overview} presents statistics for the training, development, and test sets in our corpus. ViNewsQA consists of 22,057 question-answer pairs based on 4,416 online news articles in health domain. Table \ref{tab:overview} shows the number of articles and the average lengths\footnote[6]{We use the pyvi library https://pypi.org/project/pyvi/ for word segmentation to calculate the average lengths of articles, questions and answers, and the vocabulary size.} for questions and answers, and the vocabulary size. The number of questions of our corpus is approximately equal to the number of questions of UIT-ViQuAD.

\begin{table}[h]
\caption{Overview statistics of ViNewsQA. * indicates that UIT-ViQuAD used passages as reading texts.}
\label{tab:overview}
\setlength\arrayrulewidth{1pt}
\begin{tabular}{lrrrrr}
\hline
\textbf{}                      & \textbf{Train} & \textbf{Dev} & \textbf{Test} & \textbf{All} & \textbf{UIT-ViQuAD} \\ \hline
\textbf{Number of article}              & 3,517          & 500          & 399           & {\bf 4,416}       &174 \\ 
\textbf{Number of passage*}              & -          & -          & -           & -       & 5,109 \\ 
\textbf{Number of questions}          & 17,568         & 2,497        & 1,992         & {\bf 22,057}       &23,074\\ 
\textbf{Average reading-text length}          & 342,9         & 323.9       & 360.4        & 342.4       &153.4\\ 
\textbf{Average question length}      & 10.6          & 10.8        & 10.3         & 10.6       & 12.2\\ 
\textbf{Average answer length} & 10.7          & 10.3        & 10.9        & 10.7        &8.2\\ 
\textbf{Vocabulary size}              & 29,111         & 10,765       & 10,020        & 32,749       & 41,773\\ \hline
\end{tabular}
\end{table}

\subsection{Vocabulary-based analysis}
\label{vocab}
To understand the health domain, we utilize the word cloud tool\footnote{Word cloud tool: https://www.wordclouds.com} to create graphical representations of word frequency for articles (see in Figure \ref{fig:article}), questions (see in Figure \ref{fig:question}) and answers (see in Figure \ref{fig:answer}) in our corpus. The larger the word in the visual, the more common the word is in the articles, questions, and answers. Vietnamese stop words\footnote{Vietnamese stop words: https://github.com/stopwords/vietnamese-stopwords} and numbers are excluded from these statistics. Table  \ref{tab:article}, Table \ref{tab:question} and Table \ref{tab:answer} show  the  top tenth popular words  that appears in articles, questions and answers in our corpus, respectively. These words are in the health domain, which is also characteristic of our corpus. Because the corpus is collected from the health online news articles. Five different words such as {\it bệnh nhân (patient), bác sĩ (doctor), bệnh (disease), bệnh viện (hospital), ung thư (cancer)} are appeared all in articles, questions and answers. Figure \ref{tab:UIT-ViNewsQAWordCloud} and Figure \ref{tab:ViQuADWordCloud} present word distribution for ViNewsQA and UIT-ViQuAD, respectively. Top 10 words from ViNewsQA (see Table \ref{tab:UIT-ViNewsQAWordCloud}) and UIT-ViQuAD (see Table \ref{tab:ViQuADWordCloud}) are very different because these high-frequency words on the UIT-ViQuAD corpus belong to multiple domains such as history, geography, economics, and politics.

\begin{minipage}{\textwidth}
  \begin{minipage}[b]{0.49\textwidth}
    \centering
     \includegraphics[scale=0.2]{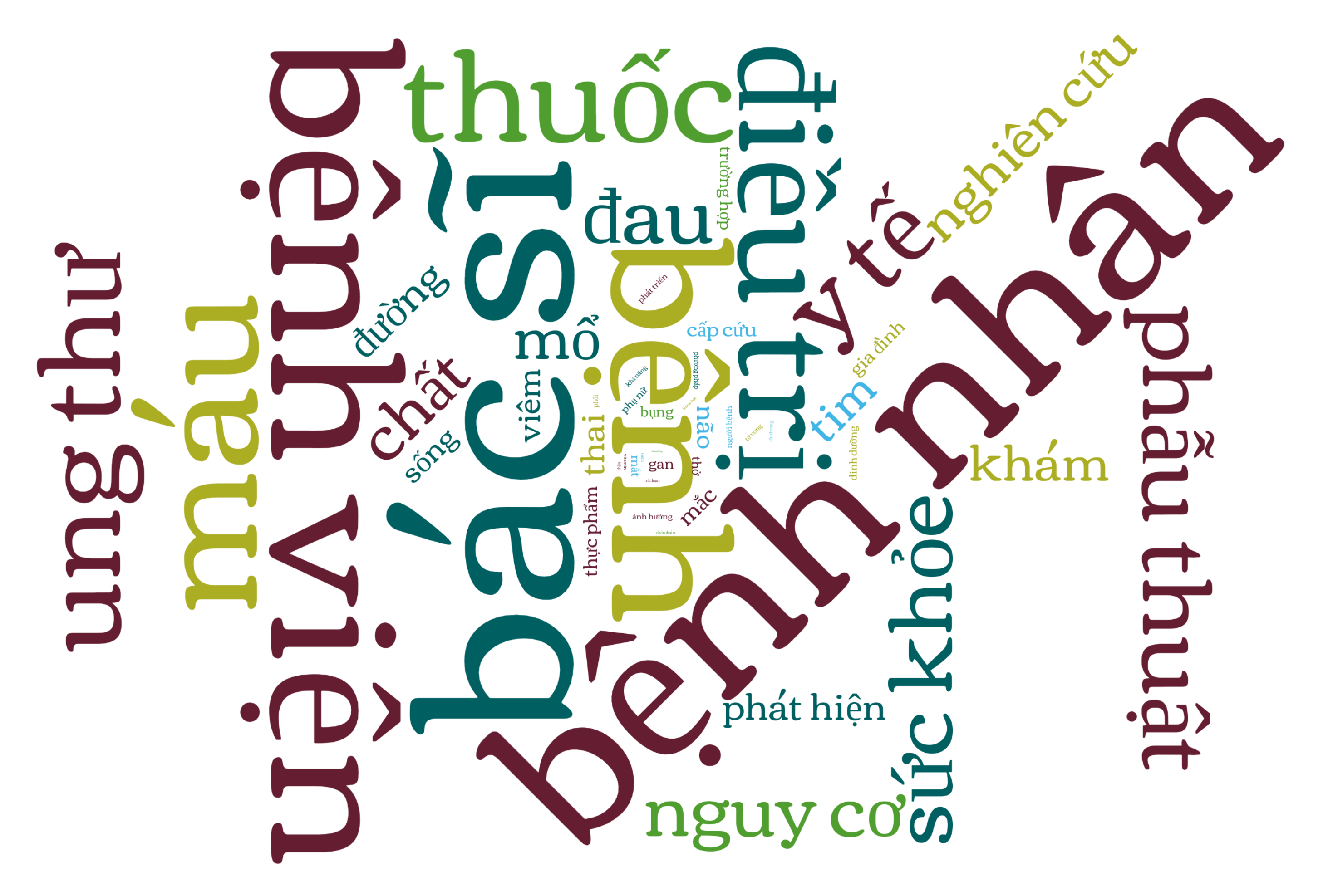} 
    \captionof{figure}{Word distribution of articles.}\label{fig:article}
  \end{minipage}
  \hfill
  \begin{minipage}[b]{0.49\textwidth}
    \centering
    \begin{tabular}{cllr}
\hline
\textbf{No.} & \multicolumn{1}{c}{\textbf{Vietnamese}} & \multicolumn{1}{c}{\textbf{English}} & \multicolumn{1}{c}{\textbf{Freq.}} \\ \hline
1 & bác sĩ & doctor & 9,314 \\ 
2 & bệnh nhân & patient & 7,182 \\ 
3 & bệnh viện  & hospital & 6,231 \\ 
4 & bệnh   & disease & 5,762 \\ 
5 & máu   & blood & 3,965 \\ 
6 & điều trị & treat & 3,901 \\
7 & thuốc  & medicine & 3,587 \\ 
8 & ung thư   & cancer & 3,205 \\ 
9 & y tế & medical & 3,066 \\ 
10 & phẫu thuật & surgery & 2,962 \\ \hline
\end{tabular}
    \captionof{table}{Common words appeared in articles.}\label{tab:article}  
    \end{minipage}
    
    \begin{minipage}[b]{0.49\textwidth}
    \centering
   
     \includegraphics[scale=0.2]{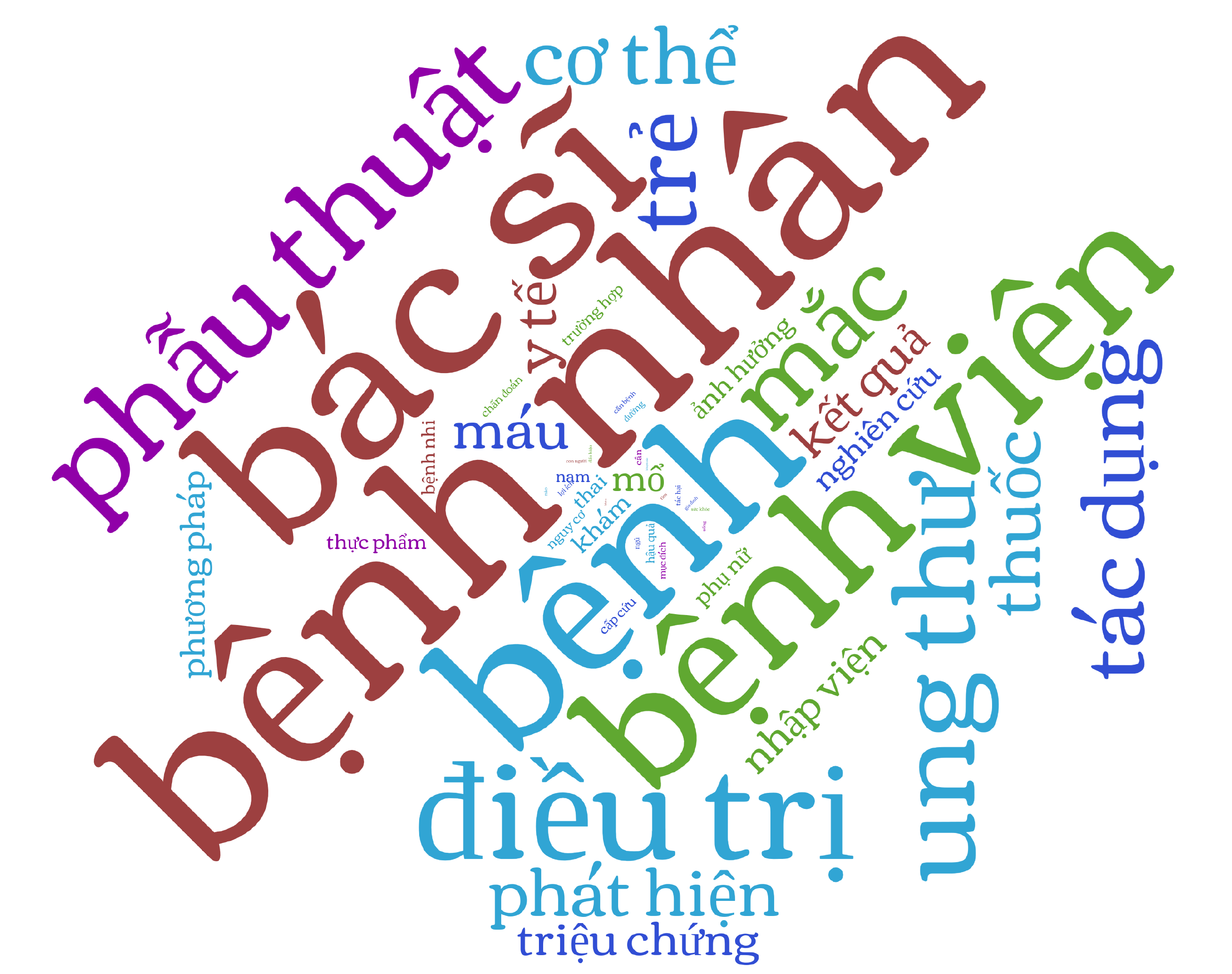} 
    \captionof{figure}{Word distribution of questions.}\label{fig:question}

  \end{minipage}
  \hfill
  \begin{minipage}[b]{0.49\textwidth}
    \centering
    \begin{tabular}{cllr}
\hline
\textbf{No.} & \multicolumn{1}{c}{\textbf{Vietnamese}} & \multicolumn{1}{c}{\textbf{English}} & \multicolumn{1}{c}{\textbf{Freq.}} \\ \hline
1 & bệnh nhân & patient & 2,281 \\ 
2 & bác sĩ & doctor & 1,753 \\ 
3 & bệnh & disease  & 1,275 \\ 
4 & bệnh viện & hospital & 1,125 \\ 
5 & nguyên nhân & reason & 694  \\ 
6 & điều trị & treat & 618 \\ 
7 & phẫu thuật & surgery & 601 \\ 
8 & ung thư & cancer & 553 \\ 
9 & mắc & get sick & 550 \\ 
10 & trẻ & young & 533  \\ \hline
\end{tabular}
    \captionof{table}{Common words appeared in questions.}\label{tab:question}
    \end{minipage}
    
    \begin{minipage}[b]{0.49\textwidth}
    \centering
     \includegraphics[scale=0.2]{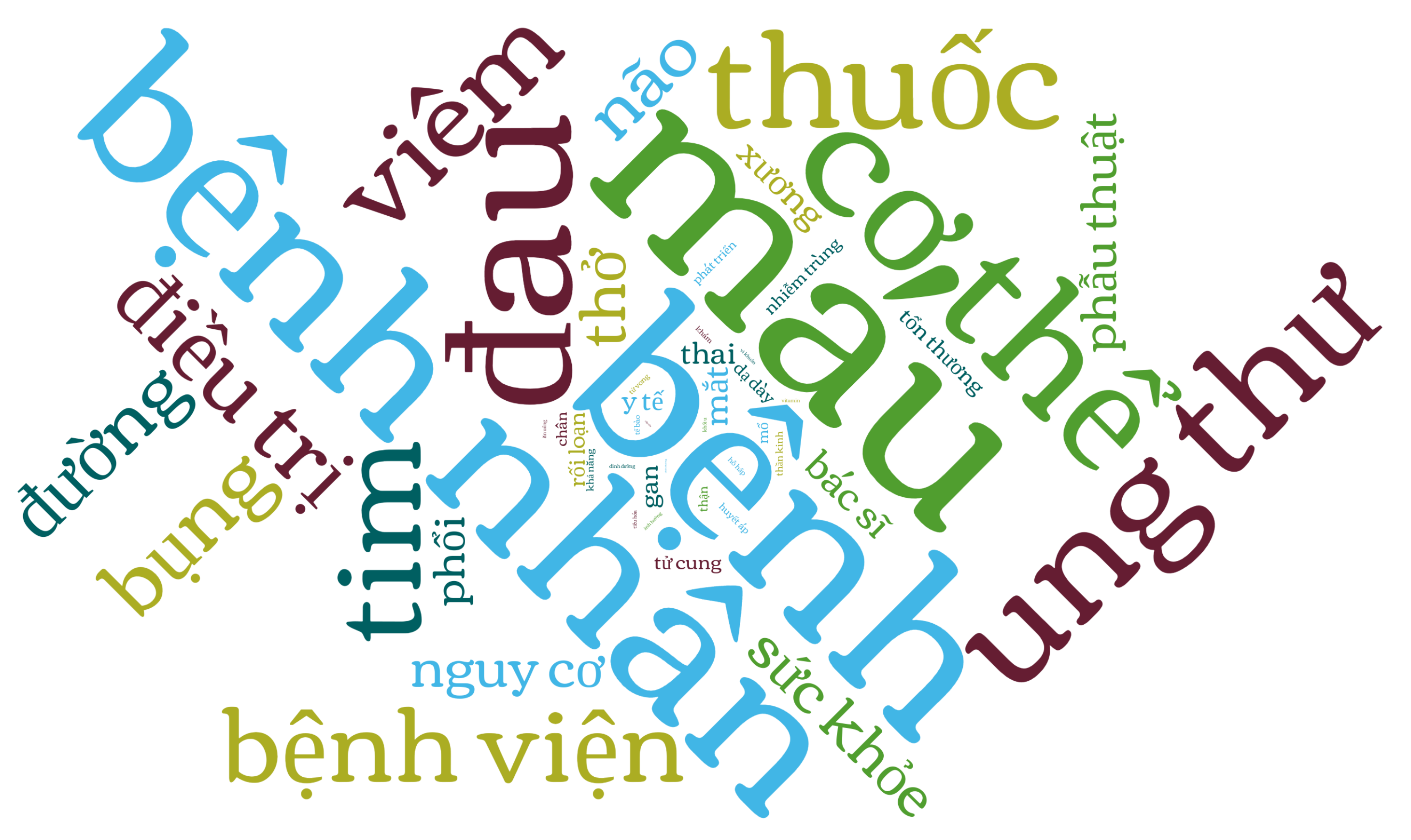} 
    \captionof{figure}{Word distribution of answers.}\label{fig:answer}
  \end{minipage}
  \hfill
  \begin{minipage}[b]{0.49\textwidth}
    \centering
    \begin{tabular}{cllr}
\hline
\textbf{No.} & \multicolumn{1}{c}{\textbf{Vietnamese}} & \multicolumn{1}{c}{\textbf{English}} & \multicolumn{1}{c}{\textbf{Freq.}} \\ \hline
1   & máu  & blood  & 2,280 \\ 
2   & bệnh & disease & 2,092 \\ 
3   & bệnh nhân & patient & 1,807 \\ 
4   & đau & pain & 1,689 \\ 
5   & cơ thể & body & 1,542 \\ 
6   & ung thư  & cancer & 1,450 \\ 
7   & thuốc & medicine  & 1,390 \\ 
8   & tim   & heart & 1,207 \\ 
9   & viêm  & inflame  & 1,143  \\ 
10  & bệnh viện & hospital  & 1,130 \\ \hline
\end{tabular}
    \captionof{table}{Common words appeared in answers.}\label{tab:answer}
    \end{minipage}
  \end{minipage}

\begin{minipage}{\textwidth}
  \begin{minipage}[b]{0.49\textwidth}
    \centering
     \includegraphics[scale=0.25]{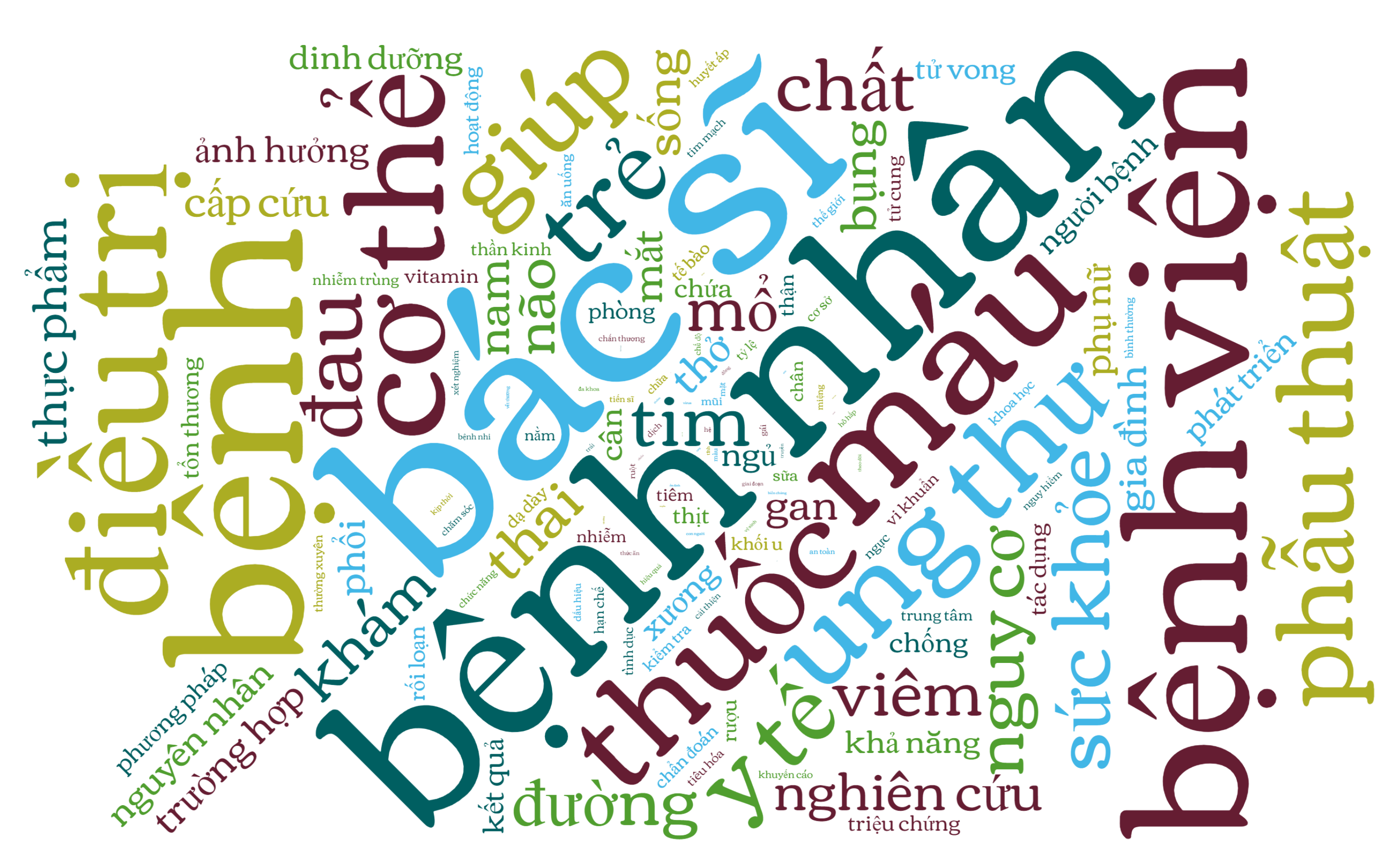} 
    \captionof{figure}{Word distribution of ViNewsQA.}\label{fig:UIT-ViNewsQAWordCloud}
  \end{minipage}
  \hfill
  \begin{minipage}[b]{0.49\textwidth}
    \centering
    \begin{tabular}{cllr}
\hline
\textbf{No.} & \multicolumn{1}{c}{\textbf{Vietnamese}} & \multicolumn{1}{c}{\textbf{English}} & \multicolumn{1}{c}{\textbf{Freq.}} \\ \hline
1 & bác sĩ & doctor & 11,628 \\ 
2 & bệnh nhân & patient & 10,878 \\ 
3 & bệnh  & disease & 8,738 \\ 
4 & bệnh viện   & hospital & 8,288 \\ 
5 & máu   & blood & 6,173 \\ 
6 & điều trị & treat & 5,393 \\
7 & cơ thể  & body & 5,173 \\ 
8 & thuốc   & medicine & 5,136 \\ 
9 & ung thư & cancer & 4,878 \\ 
10 & phẫu thuật & surgery & 4,167 \\ \hline
\end{tabular}
    \captionof{table}{Common words appeared in ViNewsQA.}\label{tab:UIT-ViNewsQAWordCloud}  
    \end{minipage}
    
    \begin{minipage}[b]{0.49\textwidth}
    \centering
   
     \includegraphics[scale=0.25]{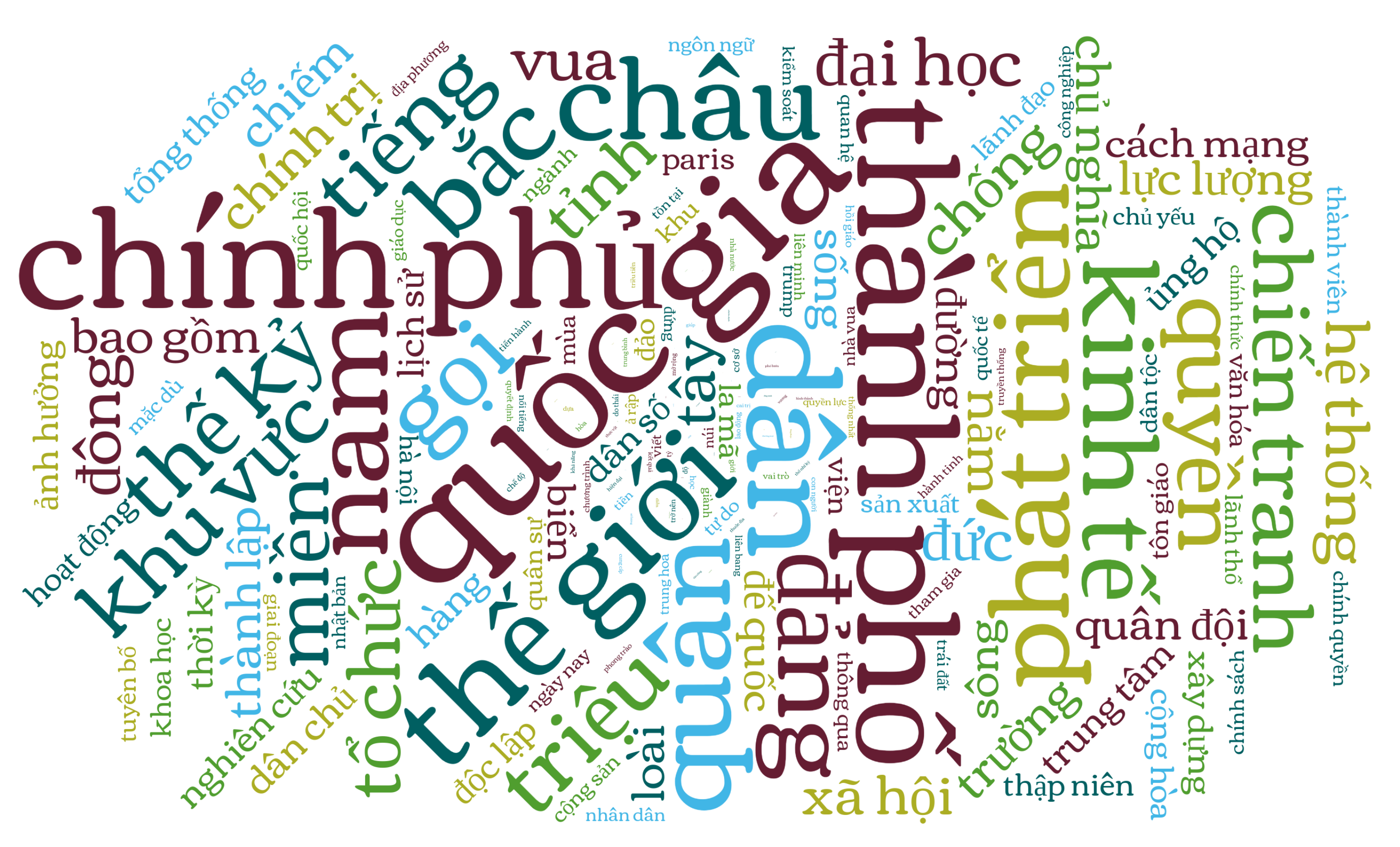} 
    \captionof{figure}{Word distribution of UIT-ViQuAD.}\label{fig:ViQuADWordCloud}

  \end{minipage}
  \hfill
  \begin{minipage}[b]{0.49\textwidth}
    \centering
    \begin{tabular}{cllr}
\hline
\textbf{No.} & \multicolumn{1}{c}{\textbf{Vietnamese}} & \multicolumn{1}{c}{\textbf{English}} & \multicolumn{1}{c}{\textbf{Freq.}} \\ \hline
1 & quốc gia & nation & 1,444 \\ 
2 & thành phố & city  & 1,264 \\ 
3 & chính phủ & goverment & 1,256 \\ 
4 & dân & resident & 1,189  \\ 
5 & thế giới & world & 1,142 \\ 
6 & nam & south & 1,077 \\ 
7 & châu & continent & 1,068  \\ 
8 & phát triển & develope & 1,045 \\ 
9 & đảng & party &  1,005\\ 
10 & kinh tế & economy & 999 \\ \hline
\end{tabular}
    \captionof{table}{Common words appeared in UIT-ViQuAD.}\label{tab:ViQuADWordCloud}
    \end{minipage}
    
  \end{minipage}

\subsection{Analysis based on different lengths}
\label{lengthana}
\subsubsection{Analysis based on question length}
Statistics for the various question lengths are shown in Table \ref{tab:qalength}. Questions with 8--9 words comprise the highest proportion with 25.67\%. Most of the questions in the corpus have lengths from 6 to 13 words, which account for approximately 80\% of the corpus. Very short questions (4--5 words) and long questions (>=18 words) account for a low percentages of 2.68\% and 3.91\%, respectively. 



                                 
        
    
       
       



\begin{table}[H]
\centering
\caption{Statistics for the question lengths in ViNewsQA.}
\label{tab:qalength}
\setlength\arrayrulewidth{1pt}
\begin{tabular}{crrrrr}
\hline

\multirow{2}{*}{{\bf Question length}} & \multicolumn{4}{c}{{\bf ViNewsQA}}                                                                           & \multirow{2}{*}{{\bf UIT-ViQuAD}} \\ \cline{2-5}
                                 & {\bf Train}                 & \multicolumn{1}{c}{{\bf Dev}} & \multicolumn{1}{c}{{\bf Test}} & \multicolumn{1}{c}{{\bf All}} &                         \\ \hline
                                 
4-5     
        
&2.73 &2.28 &2.71 &2.68 &0.99        
        \\
6-7    
    
&14.29  &14.14  &16.52  &14.48  &7.10        
        \\
8-9   
       
&25.81 &23.35 &27.31 &25.67 &17.34       
       \\
10-11   
       
&23.84  &23.91  &24.40  &23.90 &22.07        
       \\
12-13    
    
&15.82  &16.94  &13.76  &15.76  &20.28        
        \\
14-15   
       
&8.89 &10.25 &6.88 &8.86 &13.96       
       \\
16-17  
       
&4.66  &5.37  &4.62  &4.74 &8.91        
       \\
18-19  
       
&2.09 &2.08 &2.11 &2.09 &4.91         
       \\
\textgreater{}19    

&1.86 &1.68 &1.71 &1.82 &4.44 
       \\ \hline
\end{tabular}

\end{table}

\subsubsection{Analysis based on answer length}

Table \ref{tab:answerlength} shows the distribution of the answer length analysis in our corpus. The largest percentage (14.73\%) comprise answers with lengths of 3--4 words. Most of the answers (nearly 60\%) have lengths of 1--10 words. Longer answers (over 10 words) comprise a low proportion of our corpus.


       
        
        
        
        


\begin{table}[H]
\centering
\caption{Statistics of the answer lengths on ViNewsQA.}
\label{tab:answerlength}
\setlength\arrayrulewidth{1pt}
\begin{tabular}{crrrrrr}
\hline

\multirow{2}{*}{\bf Answer length} & \multicolumn{4}{c}{\bf ViNewsQA}                                                                           & \multirow{2}{*}{\bf UIT-ViQuAD} \\ \cline{2-5}
                                 & {\bf Train}                 & \multicolumn{1}{c}{\bf Dev} & \multicolumn{1}{c}{\bf Test} & \multicolumn{1}{c}{\bf All} &                         \\ \hline
1-2     
       
&12.53 &14.82 &13.25 &12.85 &29.46        \\
3-4    
        
&14.80 &15.18 &13.60 &14.73 &17.64        \\
5-6   
        
&11.48 &12.37 &12.75 &11.69 &12.10        \\
7-8   
        
&10.50 &10.37 &9.04 &10.35 &7.95        \\
9-10     
       
&9.91 &9.33 &8.23 &9.69 &6.39        \\
11-12    
        
&8.64 &7.81 &9.04 &8.58 &5.03        \\
13-14   
        
&7.02 &6.69 &6.83 &6.96 &4.17        \\
15-16   
        
&5.85 &5.01 &6.73 &5.83 &3.25        \\
17-18    
        
&4.99 &3.80 &5.17 &4.87 &2.62        \\
19-20   
        
&3.67 &3.16 &3.56 &3.60 &2.26        \\
\textgreater{}20    
        
&10.62 &11.45 &11.80 &10.82 &9.13       \\ \hline
\end{tabular}

\end{table}

\subsubsection{Analysis based on article length}
\begin{figure}[H]
\centering
\includegraphics[width=14cm]{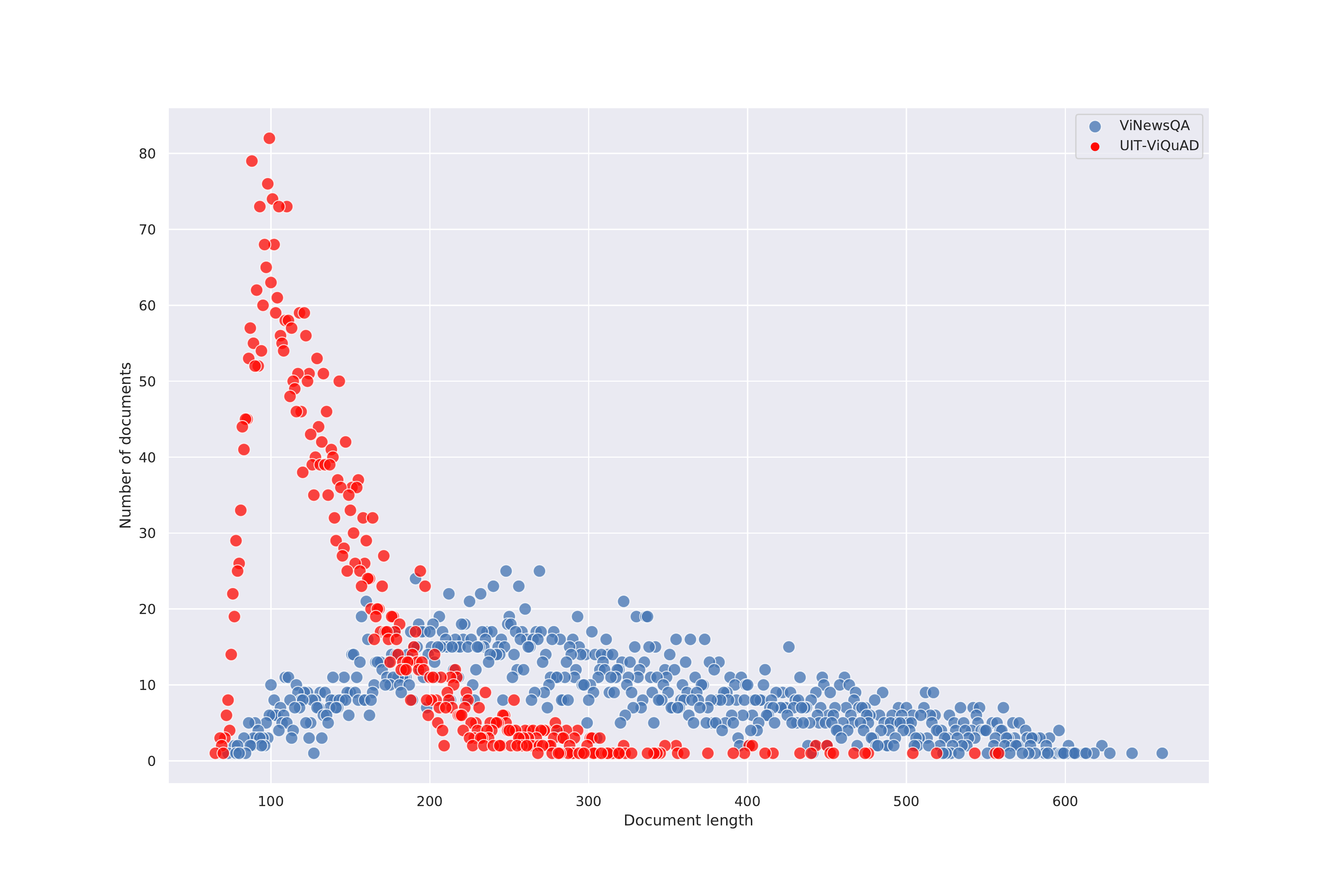}
\caption{Document length distributions of two corpora (ViNewsQA and UIT-ViQuAD).}\label{fig:len-distribution}
\centering
\end{figure}

Besides, we also aim to analyze article lengths in the corpus. Table \ref{tab:readingtextlength} presents statistics for various article lengths in our corpus. The lengths of most articles ranging from 101 to 500 words, which account for over 84\%. The length of the reading texts on ViNewsQA is significantly longer than that on the UIT-ViQuAD. Figure \ref{fig:len-distribution} shows different reading-text-length distributions of the two Vietnamese corpora. Based on the length characteristics, we determine whether the length affects the performance of the machine models and humans according to question, answer, or article lengths?



  

 

    


\begin{table}[H]
\centering
\caption{Statistics for ViNewsQA according to the article length.}
\label{tab:readingtextlength}
\setlength\arrayrulewidth{1pt}
\begin{tabular}{crrrr}
\hline
{\bf Article length} 
                                 & {\bf Train}                 & \multicolumn{1}{c}{\bf Dev} & \multicolumn{1}{c}{\bf Test} & \multicolumn{1}{c}{\bf All}                         \\ \hline
\textless{}101      
&0.34 &0.40 &0.25 &0.34                          \\

101-200             
  
&15.33 &18.40 &13.53 &15.51                         \\

201-300             
 
&29.03 &31.40 &27.07 &29.12                          \\
301-400             
  
&23.34 &23.60 &23.56 &23.39                            \\

401-500             
 
&16.52 &13.40 &16.29 &16.15                        \\
501-600             
 
&10.41 &7.80 &11.03 &10.17                         \\
\textgreater{}600   
    
&5.03 &5.00 &8.27 &5.32 \\ \hline
\end{tabular}

\end{table}

\subsection{Analysis based on different types}
\label{typeana}

Before conducting the type-based analysis, we hire three annotators and train them on 100 questions to have more than 80\% of the Cohen's kappa inter-annotator agreement before annotating data simultaneously. 
\subsubsection{Analysis based on question type}
In this work, we divide Vietnamese questions into seven different question types such as Who, What, When, Where, Why, How, and Others, which is constructed in a similar way to UIT-ViQuAD \cite{Kiet:20ViQuAD} and CMRC \cite{Cui:18}. These question types are described as follows. 

\begin{itemize}
    \item {\bf Who:} A group of questions have their answers related to people. 
    \item {\bf When:} A group of questions requires their answers presenting time expression.
    \item {\bf Where:} A group of questions requires their answers which are locations or places.
    \item {\bf Why:} A group of questions require their answers expressing reasons.
    \item {\bf How:} A group of questions require their answers related to an away or method to do.
    \item {\bf What:} A group of questions have answers which are definitions, things or events.
    \item {\bf Others:} A group of questions do not belong to the above types. Most of the questions are related to numbers such How many and How much.
\end{itemize}

Table \ref{tab:questiontype1} presents the distribution of the question types on our corpus.  The table show that the question type What accounted for the largest proportion with 54.35\%. Compared to the SQuAD corpus, the rate of the What question in our corpus is similar to that in SQuAD (49.97\%) \cite{aniol2019ensemble}. Our corpus requires abilities beyond factoid questions that demand intricate knowledge and skills to answer like Why and How questions. In particular, How and Why ranked the second and the third with 13.46\% and 12.17\%, respectively.

\begin{table}[H]
\caption{Statistics and examples of question type.}
\label{tab:questiontype1}
\setlength\arrayrulewidth{1pt}
\begin{tabular}{p{1cm}p{7cm}p{2cm}p{2cm}}
\hline
\multirow{2}{*}{{\bf Type}}  & \multirow{2}{*}{{\bf Example}} & \multicolumn{2}{c}{{\bf Percentage (\%)}}                \\ \cline{3-4} 
                       &                          & {\bf ViNewsQA}              & \multicolumn{1}{c}{{\bf UIT-ViQuAD}} \\ \hline
Who &
  {\bf Vietnamese}: Bệnh viện Thẩm mỹ Emcas đã hợp tác với ai để thực hiện ca phẫu thuật? 
  {\bf English}: Who has Emcas Cosmetic Hospital cooperated with to perform the surgery? & \multicolumn{1}{r}{3.16} &\multicolumn{1}{r}{9.41}\\  \hline
When &
  {\bf Vietnamese}: Mỹ dùng năng lượng vi sóng để điều trị hôi nách vào năm nào?
  
  {\bf English}: When did the US use microwave energy to treat armpits? & \multicolumn{1}{r}{3.52} &\multicolumn{1}{r}{8.96}\\ \hline
Where &
  {\bf Vietnamese}: Bệnh viện đã cử 14 người đi đào tạo ở đâu? 
  
  {\bf English}: Where did the hospital take 14 people to train? & \multicolumn{1}{r}{3.40} &\multicolumn{1}{r}{5.64}\\ \hline 
Why &
  {\bf Vietnamese}: Tại sao phụ nữ ở Mỹ ngày càng sinh con muộn?
  
  {\bf English}: Why are women in America increasingly late for giving birth? & \multicolumn{1}{r}{12.17} &\multicolumn{1}{r}{7.54}\\ \hline
How &
  {\bf Vietnamese}: Bệnh viêm não cấp thường lây nhiễm bằng cách nào?
  
  {\bf English}: How is acute encephalitis usually spread? & \multicolumn{1}{r}{13.46} &\multicolumn{1}{r}{9.09}\\ \hline
What &
  {\bf Vietnamese}: Trong các thức uống năng lượng có những thành phần nào giống nhau?
  
  {\bf English}: What are the similar ingredients in energy drinks? &\multicolumn{1}{r}{54.35} &\multicolumn{1}{r}{49.97}\\ \hline 
Others &
  {\bf Vietnamese}: Hà nội đã tiếp nhận bao nhiêu đơn vị máu từ người hiến tình nguyện vào năm 2018? 
  
  {\bf English}: How many units of blood did Hanoi receive from voluntary donors in 2018? & \multicolumn{1}{r}{9.94} &\multicolumn{1}{r}{9.41}\\ \hline
\end{tabular}
\end{table}

\subsubsection{Analysis based on answer type}
We divide answers into 11 types including numbers (time, other numeric), entity (person, location, other entity), phrase (noun phrase, adjective phrase, verb phrase, prepositional phrase, clause) and others. The priority order of annotation is number, entity, phrase and others. Table \ref{tab:answertype} present statistics of answer types in our corpus. While verb phrases account for the highest proportion of 34.84\%, prepositional phrases account for the lowest proportion with 0.8\%.

\begin{longtable}[c]{p{1.5cm}p{6.5cm}rr}
\caption{Statistics and examples of answer type.}
\label{tab:answertype}\\
\hline
\multicolumn{1}{c}{\multirow{2}{*}{\textbf{Type}}} & \multicolumn{1}{c}{\multirow{2}{*}{\textbf{Example}}}                                                                                                                                                                                                            & \multicolumn{2}{c}{\textbf{Percentage (\%)}}                                     \\ \cline{3-4} 
\multicolumn{1}{c}{}                               & \multicolumn{1}{c}{}                                                                                                                                                                                                                                             & \multicolumn{1}{c}{\textbf{ViNewsQA}} & \multicolumn{1}{c}{\textbf{UIT-ViQuAD}} \\ \hline
\endfirsthead
\endhead
Time                                                 & {\bf Vietnamese}: tháng 5/2017, tuần thứ 36.   \\ &{\bf English}: May 2017, 36th week.                                                                                & 4.49                                   & 7.71                                     \\ \hline
Other numeric                                        & {\bf Vietnamese}: 12\%, hơn 350 calo.\\   &{\bf English}: 12\%, more than 350 calories.                                                                           & 9.29                                   & 9.41                                     \\ \hline
Person                                               & {\bf Vietnamese}: Bác sĩ Nguyễn Khắc Vui, nhiếp ảnh gia Amy Taylor.\\ \\  &{\bf English}: Dr. Nguyen Khac Vui, photographer Amy Taylor.                      & 0.96                                   & 5.39                                     \\ \hline
Location                                             & {\bf Vietnamese}: bệnh viện Westchester, Quận 8.\\ &{\bf English}: Westchester Hospital, District 8. & 1.16                                   & 4.32                                     \\ \hline
Other entity                                         & {\bf Vietnamese}: Cục An toàn thực phẩm, khoa Hồi sức cấp cứu.   \\ &{\bf English}: Food Safety Department, Emergency Care Department.                                                                           & 4.25                                   & 11.65                                    \\ \hline
Noun phrase                                          & {\bf Vietnamese}: trẻ em, sự lây lan của vi khuẩn.   \\ &{\bf English}: children, spread of bacteria.                                                                          & 27.63                                  & 22.86                                    \\ \hline
Adjective phrase                                     & {\bf Vietnamese}: rất đắt đỏ, béo phì.   \\ &{\bf English}: very expensive, fat.                                                                                             & 4.41                                   & 2.52                                     \\ \hline
Verb phrase                                          & {\bf Vietnamese}: kiểm tra huyết áp, uống một ly nước chanh. \\   \\  &{\bf English}: check blood pressure, drink a glass of lemon juice.         &34.84                                  & 18.43                                    \\ \hline
Preposition phrase                                   & {\bf Vietnamese}: dưới gan phải, ở vùng đáy tử cung.    \\  &{\bf English}: under the right liver, in the base of the uterus.                                       & 0.80                                   & 3.18                                     \\ \hline
Clause                                               & {\bf Vietnamese}: mỗi tình nguyện viên sẽ cần uống khoảng 1.000 chai rượu mỗi ngày, Thuốc lá làm cạn lượng vitamin C trong cơ thể.\\  &{\bf English}: Each volunteer will need to drink about 1,000 bottles of wine per day, Tobacco depletes vitamin C in the body.                                                                                                          & 5.65                                   & 5.91                                     \\ \hline
Others                                               & {\bf Vietnamese}:  Tôi đã sống một cuộc đời rất hạnh phúc và viên mãn nên không còn gì phải hối hận. Không quan trọng là mình sống được bao lâu, quan trọng là ý nghĩa mỗi ngày được sống.   \\  &{\bf English}: I have lived a very happy and fulfilled life, so I have no regrets. It doesn't matter how long we live, what matters to live each day.                                                                                                                                             & 6.52                                   & 10.55                                    \\ \hline
\end{longtable}

\subsubsection{Analysis based on reasoning type}
To classify the difficulty of a question, we divided question reasoning into one of five types, comprising word matching, paraphrasing, single-sentence reasoning, multi-sentence reasoning, and ambiguous/insufficient. These reasoning types are described as follows.

\begin{itemize}
    \item {\bf Word matching} (WM): The main words in the question exactly match the words in the reading text.
    \item {\bf Paraphrasing} (PP): The questions are paraphrased from a single sentence in the reading text. In particular, we may use synonymy and world knowledge to create the question.
    \item {\bf Single-sentence Reasoning} (SSR): The answers are inferred from a single sentence in the article. Such answers could be created by extracting incomplete information or conceptual overlap.
    \item {\bf Multi-sentence Reasoning} (MSR): The answers are inferred from multiple sentences in the article by information fusion techniques.
    \item {\bf Ambiguous/Insufficient} (AoI):  The questions have many answers or answers are not found in the article.
\end{itemize}

The reasoning types in the development set for our corpus annotated by workers. Table \ref{tab:reasonType} shows the distributions of the reasoning types in our corpus. The reasoning of the largest proportion is paraphrasing (PP) with 31.16\%, whereas the lowest is Ambiguous/Insufficient (AoI) with 0.40\%.

\begin{longtable}[c]{p{1.5cm}p{7cm}p{2cm}p{2cm}}
\caption{Statistics and examples of reasoning type.}
\label{tab:reasonType}
\\
\hline
\multirow{2}{*}{{\bf Reasoning}}  & \multirow{2}{*}{{\bf Example}} & \multicolumn{2}{c}{{\bf Percentage (\%)}}                \\ \cline{3-4} 
                       &                          & {\bf ViNewsQA}              & \multicolumn{1}{c}{{\bf UIT-ViQuAD}} \\ \hline

Word matching &
    \begin{tabular}[c]{p{7cm}}\textbf{\underline{Context:}} Thành phần trong nhân sâm chứa nhiều \textbf{\textit{ginsenoside}} giúp cải thiện và tăng đáng kể hàm lượng testosterone ở nam giới, làm tăng cảm giác hưng phấn.\\ \textbf{\underline{Question:}} Thành phần trong nhân sâm chứa nhiều chất gì? \\ \textbf{\underline{Answer:}} \textbf{\textit{ginsenoside}} \\ \\ \textbf{English translation:} \\ \textbf{\underline{Context:}} The ingredient in ginseng contains \textbf{\textit{ginsenoside.}} which helps improve and significantly increase testosterone content in men, increasing feelings of excitement. \\ \textbf{\underline{Question:}} What are the ingredients in ginseng? \\ \textbf{\underline{Answer:}} \textbf{\textit{ginsenoside.}}\end{tabular} & \multicolumn{1}{r}{26.19}   &\multicolumn{1}{r}{13.35}\\ \hline 
 
Paraphrasing & 
    \begin{tabular}[c]{p{7cm}}\textbf{\underline{Context:}} Bác sĩ Nguyễn Hữu Thịnh cho biết người bị tắc ruột thường có các triệu chứng \textbf{\textit{buồn nôn, trướng bụng, không thể đại tiện hoặc trung tiện, đau bụng quặn từng cơn.}} .\\ \textbf{\underline{Question:}} Người bị tắc ruột thường có các dấu hiệu gì? \\ \textbf{\underline{Answer:}} \textbf{\textit{buồn nôn, trướng bụng, không thể đại tiện hoặc trung tiện, đau bụng quặn từng cơn.}} \\\\ \textbf{English translation:} \\ \textbf{\underline{Context:}} Doctor Nguyen Huu Thinh said people with intestinal obstruction often have symptoms of \textbf{\textit{nausea, abdominal distention, unable to defecate or defecate, abdominal pain with intermittent cramps.}}. \\ \textbf{\underline{Question:}} What are the indications of a person with intestinal obstruction? \\ \textbf{\underline{Answer:}} \textbf{\textit{nausea, abdominal distention, unable to defecate or defecate, abdominal pain with intermittent cramps.}}\end{tabular} 
    & \multicolumn{1}{r}{31.16}  &\multicolumn{1}{r}{31.22}\\ \hline

Single-sentence reasoning & 
    \begin{tabular}[c]{p{7cm}}\textbf{\underline{Context:}} Hội chứng người sói Congenital Hypertrichosis hay hội chứng người sói là căn bệnh di truyền hiếm gặp khiến \textbf{\textit{lông, tóc trên toàn bộ cơ thể phát triển quá mức}}.\\ \textbf{\underline{Question:}} Người bị hội chứng người sói sẽ có những điểm gì khác biệt?\\ \textbf{\underline{Answer:}} \textbf{\textit{lông, tóc trên toàn bộ cơ thể phát triển quá mức.}}\\\\ \textbf{English translation:}\\ \textbf{\underline{Context:}} The werewolf syndrome Congenital Hypertrichosis or werewolf syndrome is a rare genetic disease that causes \textbf{\textit{excessive hair and hair growth throughout the body.}}. \\ \textbf{\underline{Question:}} What are the differences between people with werewolf syndrome? \\ \textbf{\underline{Answer:}} \textbf{\textit{excessive hair and hair growth throughout the body.}}\end{tabular} 
    & \multicolumn{1}{r}{16.78} &\multicolumn{1}{r}{38.07}\\ \hline

Multiple-sentence reasoning & 
    \begin{tabular}[c]{p{7cm}}\textbf{\underline{Context:}} Theo Bảng thành phần dinh dưỡng Việt Nam, \textbf{\textit{trứng gà ít calo và cholesterol hơn trứng vịt}}. Bác sĩ khuyên mỗi ngày chỉ nên ăn một quả trứng gà  thay vì trứng vịt sẽ phù hợp với những người huyết áp cao, tim mạch. \\ \textbf{\underline{Question:}} Tại sao người huyết áp cao và tim mạch nên ăn trứng gà thay vì trứng vịt? \\ \textbf{\underline{Answer:}} \textbf{\textit{trứng gà ít calo và cholesterol hơn trứng vịt.}} \\\\ \textbf{English translation:}\\ \textbf{\underline{Context:}} According to the Vietnam Nutrition Facts Table, \textbf{\textit{chicken eggs are lower in calories and cholesterol than duck eggs}}. Doctors recommend eating only one egg per day instead of duck eggs, which will be suitable for people with high blood pressure, cardiovascular.\\ \textbf{\underline{Question:}} Why should high blood pressure and cardiovascular patients eat chicken eggs instead of duck eggs?\\ \textbf{\underline{Answer:}} \textbf{\textit{chicken eggs are lower in calories and cholesterol than duck eggs.}}\end{tabular} 
    & \multicolumn{1}{r}{25.47} &\multicolumn{1}{r}{16.22} \\ \hline

Ambiguous or insufficient & 
    \begin{tabular}[c]{p{7cm}}\textbf{\underline{Context:}} Bé được chuyển lên phòng mổ cấp cứu, bác sĩ xác định bỏng nặng độ I,II,III. Bệnh nhi \textbf{\textit{phải ghép da, vá da dày toàn bộ, băng ép cố định diện ghép da, nẹp cố định cánh, cẳng tay trái bằng nẹp bột, vệ sinh làm sạch vết thương hằng ngày… và chăm sóc đặc biệt}}\\ \textbf{\underline{Question:}} Bệnh nhi cần được cấp cứu như thế nào?\\ \textbf{\underline{Answer:}} \textbf{\textit{chăm sóc đặc biệt}} \\ \textbf{Correct:} \textbf{\textit{phải ghép da, vá da dày toàn bộ, băng ép cố định diện ghép da, nẹp cố định cánh, cẳng tay trái bằng nẹp bột, vệ sinh làm sạch vết thương hằng ngày… và chăm sóc đặt biệt}}. \\\\ \textbf{English translation:}\\ \textbf{\underline{Context:}} The baby was transferred to the emergency operating room, the doctor determined severe burns I, II, III. Patients \textbf{\textit{must have skin grafts, thick skin patches, compression bandages for fixed skin grafts, fixed braces, left forearms with powder splint, cleaning the wound daily… and special care}}.\\ \textbf{\underline{Question:}} How do children need emergency care?\\ \textbf{\underline{Answer:}} \textbf{\textit{special care}} \\ \textbf{Correct:} \textbf{\textit{must have skin grafts, thick skin patches, compression bandages for fixed skin grafts, fixed braces, left forearms with powder splint, cleaning the wound daily… and special care}}.\end{tabular} 
    & \multicolumn{1}{r}{0.40} &\multicolumn{1}{r}{2.11} \\ \hline
\end{longtable}

\subsection{Correlation between answer length and different types}
\label{correlation}
To understand relations between answer length and different linguistic types, we analyze various correlations consisting of question type-answer length, answer type-answer length, and reasoning type-answer length. Nguyen et al. \cite{Kiet:20ViQuAD} shows that answer length had an apparent effect, and the longer the answer is, the more difficult it is. Therefore, we choose answer length to analyze along with types. The specific results of our analysis are presented as follows.

Table \ref{tab:answer-questiontype} presents the correlation between question type and answer length. As can be seen from the table, answers for Who, When, and Where tend to be short and are mostly 1 to 5 words in length. In contrast, answers for Why, How, and What tend to be long and varied in length because they are more difficult than Who, When, and Where questions. Others account for a large proportion (>90\%) for 1 to 5-word answers because they are mostly related to the number, such as How many and How much.

\begin{table}[H]
\centering
\caption{The relation between question type and answer length.}
\label{tab:answer-questiontype}
\begin{tabular}{lrrrrr}
\hline
\multicolumn{1}{c}{\textbf{Answer length}} & \multicolumn{1}{c}{\textbf{1-5}} & \multicolumn{1}{c}{\textbf{6-10}} & \multicolumn{1}{c}{\textbf{11-15}} & \multicolumn{1}{c}{\textbf{15-20}} & \multicolumn{1}{c}{\textbf{\textgreater{}20}} \\ \hline
Who                                       & 51.90                             & 21.52                              & 10.13                               & 7.59                                & 8.86                                           \\
When                                       & 75.00                             & 18.18                              & 4.55                                & 1.14                                & 1.14                                           \\ 
Where                                       & 62.35                             & 17.65                              & 12.94                               & 2.35                                & 4.71                                           \\
Why                                       & 24.67                             & 32.57                              & 19.74                               & 12.50                               & 10.53                                          \\
How                                       & 28.37                             & 28.74                              & 19.60                               & 9.80                                & 13.49                                          \\ 
What                                       & 18.45                             & 29.76                              & 20.83                               & 13.99                               & 16.96                                          \\
Others                                       & 90.20                             & 6.12                               & 2.86                                & 0.41                                & 0.41                                           \\ \hline
\end{tabular}
\end{table}


Table \ref{tab:answer-answertype} shows the correlation between answer type and answer length. Answers based on numbers (time and other numeric), entities (person, location and other entity) and preposition phrases tend to be shorter than word answers based on noun phrases, verb phrases, adjective phrases, clauses and others.

\begin{table}[H]
\centering
\caption{The relation between answer type and answer length.}
\label{tab:answer-answertype}
\begin{tabular}{lrrrrr}
\hline
\multicolumn{1}{c}{\textbf{Answer length}} & \multicolumn{1}{c}{\textbf{1-5}} & \multicolumn{1}{c}{\textbf{6-10}} & \multicolumn{1}{c}{\textbf{11-15}} & \multicolumn{1}{c}{\textbf{16-20}} & \multicolumn{1}{c}{\textbf{\textgreater{}20}} \\ \hline
Time                                       & 85.71                             & 12.50                              & 0.00                                & 0.00                                & 1.79                                           \\ 
Other numeric                                       & 86.21                             & 9.48                               & 3.02                                & 0.86                                & 0.43                                           \\ 
Person                                       & 91.67                             & 8.33                               & 0.00                                & 0.00                                & 0.00                                           \\ 
Location                                       & 72.41                             & 17.24                              & 10.34                               & 0.00                                & 0.00                                           \\ 
Other entity                                       & 73.58                             & 13.21                              & 7.55                                & 0.94                                & 4.72                                           \\ 
Noun phrase                                       & 33.19                             & 30.87                              & 18.26                               & 7.83                                & 9.86                                           \\ 
Verb phrase                                        & 21.84                             & 31.15                              & 21.61                               & 13.10                               & 12.30                                          \\ 
Adjective phrase                                       & 25.45                             & 40.91                              & 19.09                               & 5.45                                & 9.09                                           \\ 
Preposition phrase                                       & 40.00                             & 25.00                              & 30.00                               & 0.00                                & 0.00                                           \\ 
Clause                                      & 13.48                             & 20.57                              & 23.40                               & 21.99                               & 20.57                                          \\ 
Others                                      & 7.89                              & 20.86                              & 20.86                               & 12.27                               & 38.04                                          \\ \hline
\end{tabular}
\end{table}

Table \ref{tab:answer-reasoningtype} shows the correlation between reasoning type and answer length. Multi-sentence reasoning questions tend to be longer than other questions. In contrast, word matching, paraphrasing and single-sentence reasoning questions have almost the same ratio. 

\begin{table}[H]
\centering
\caption{The relation between reasoning type and answer length.}
\label{tab:answer-reasoningtype}
\begin{tabular}{llllll}
\hline
\multicolumn{1}{c}{\textbf{Answer length}} & \multicolumn{1}{c}{\textbf{1-5}} & \multicolumn{1}{c}{\textbf{6-10}} & \multicolumn{1}{c}{\textbf{11-15}} & \multicolumn{1}{c}{\textbf{16-20}} & \multicolumn{1}{c}{\textbf{\textgreater{}20}} \\ \hline
Word matching                                & 41.74                             & 26.61                              & 15.29                               & 7.34                                & 9.02                                           \\ 
Paraphrasing                                 & 36.63                             & 27.51                              & 17.87                               & 8.35                                & 9.64                                           \\ 
Single-sentence reasoning                    & 40.57                             & 27.21                              & 14.80                               & 8.11                                & 9.31                                           \\ 
Multi-sentence resoning                      & 26.89                             & 23.43                              & 19.50                               & 12.74                               & 17.45                                          \\ 
Ambiguous or insufficient                    & 50.00                             & 30.00                              & 0.00                                & 0.00                                & 10.00                                          \\ \hline
\end{tabular}
\end{table}

\section{Empirical Evaluation}
\label{experiment}
In this section, we aim to evaluate three different types of methods consisting of rule-based, neural network-based, and transfer learning-based models on our corpus. Our experiments revolve around the following questions: {\bf Q1}: How do the MRC models perform on this new corpus we build? and {\bf Q2}: Do these models outperform humans? 


\subsection{Re-implemented methods and baselines}
\label{method}

To solve the research question {\bf Q1}, MRC models are chosen: rule-based (Sliding Window), neural network-based (DrQA and QANet), and transfer learning-based (BERT and ALBERT). Because these models are popular for evaluating machine reading comprehension in English \cite{Che:2017,Yu:18,Dev:18,Con:19} and in other languages \cite{Cui:18,Lim:19,Mar:20,Pav:20,Kiet:20ViQuAD}. This section conducts various experiments to evaluate the MRC methods as the first baseline models on our corpus. 


    


\begin{figure}[H]
\includegraphics[width=12cm]{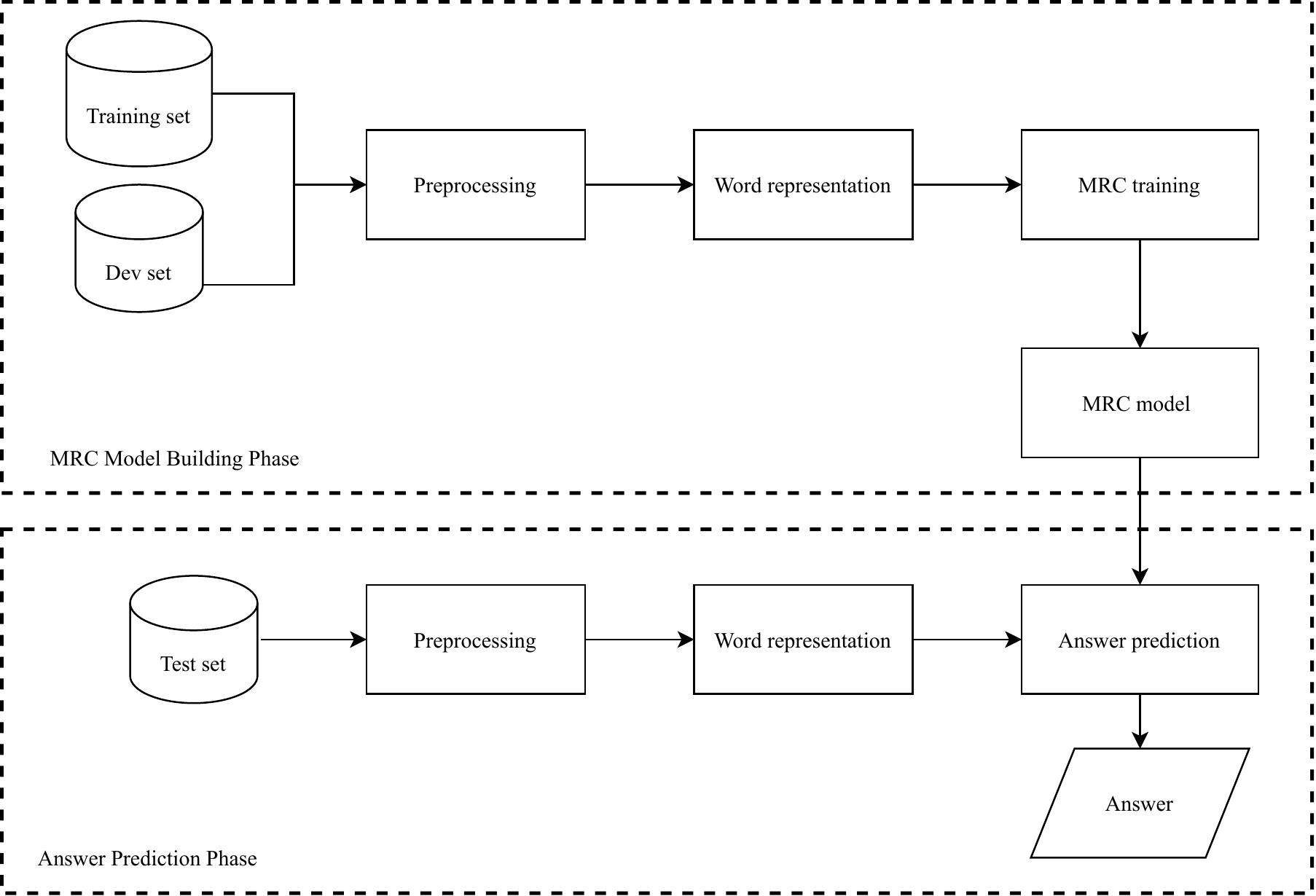}
\caption{Baseline framework for machine reading comprehension in Vietnamese.}\label{fig:OverallModel}
\centering
\end{figure}

Sliding Window is the first baseline model to conduct experiments on the well-known different corpora such as MCTest \cite{Ric:13}, SQuAD \cite{Raj:16}, DREAM \cite{Sun:19}, and ViMMRC \cite{Ngu:20}. In recent years, neural network-based and transfer learning-based models have achieved better performances on the span-extraction MRC corpora. How do these methods work well on our corpus? Hence, we aim to evaluate these methods as baseline models on our corpus. To adapt these methods into our corpus, we follow the Vietnamese MRC framework, as shown in Figure \ref{fig:OverallModel}. The pre-processing stage before training involved word segmentation, removing extra white spaces, and updating the answer positions. In addition, we remove punctuations when performing evaluations. We use pre-trained embeddings for word representations, which have proved effective in the MRC task in Vietnamese \cite{Kiet:20ViQuAD,Ngu:20}. Four popular neural network-based and transfer learning-based MRC methods are chosen for the framework: DrQA Reader \cite{Che:2017}, QANet \cite{Yu:18}, BERT \cite{Dev:18} and ALBERT \cite{lan2019albert}, because these methods have achieved state-of-the-art performances in many MRC tasks on SQuAD \cite{Dev:18}, NewsQA \cite{joshi2020spanbert}, and CMRC \cite{Cui:18}. These models are described as follows.

\begin{itemize}

\item {\bf Sliding Window}: Sliding window (SW) is the first rule-based approach proposed by Richardson et al. (2013) \cite{Ric:13}. This approach matches a set of words built from a question and one of its answer candidates with a given document before calculating the matching score using TF-IDF for each answer candidate. Experiments have been conducted with this simple model on many different corpora as first baseline models, such as MCTest \cite{Ric:13}, SQuAD \cite{Raj:16}, DREAM \cite{Sun:19}, and ViMMRC \cite{Ngu:20}.

\item {\bf DrQA}: We implement the simple but effective neural-based model DrQA Reader, which is based on the open-domain QA system called DrQA proposed by Chen et al. \cite{Che:2017} in 2017. This model has achieved good performance with multiple MRC corpora such as SQuAD  \cite{Raj:16} and CoQA \cite{Red:19}. DrQA is known as a simple-reasoning MRC model with multiple layers. In the input layer, the model presents binary features to the lexical-unit embedding of each context lexical unit if this lexical unit and its variants appear in the question. 
In addition, the model extends the lexical-unit embeddings in the input layer with the linguistic features such as POS and NER. The UIT-ViQuAD corpus also used DrQA as one of first baseline model \cite{Kiet:20ViQuAD}.

\item {\bf QANet}: Yu et al. proposed QANet \shortcite{Yu:18} and this model has obtained good performance with many MRC corpora \cite{Raj:16,Dua:19}. QANet has a feed-forward architecture with convolutions and attention mechanisms for MRC. The model comprises multiple convolutional layers followed by two components: the self-attention and fully connected layer, for both question and reading text encoding, as well as three stacked layers before predicting the final output. It was also the first baseline model of the UIT-ViQuAD corpus \cite{Kiet:20ViQuAD}.


\item {\bf BERT}: Devlin et al. \shortcite{Dev:18} recently proposed BERT, which uses a transformer network to pre-train a language model for extracting contextual word embeddings. This model is one of the best for contextualized representation learning \cite{pet:18,how:18,rad:18,Dev:18} and it has achieved the state-of-the-art results in multiple reading comprehension tasks. In this study, we used mBERT \cite{Dev:18}, as a large-scale multilingual language model, which was pre-trained for evaluating our Vietnamese MRC task. Nguyen et al. \cite{Kiet:20ViQuAD} chose BERT as one of the first baseline models.

\item {\bf ALBERT}: Lan et al. proposed ALBERT \cite{lan2019albert} to improve NLP tasks. This model was designed similarly to BERT, using the transformer encoder architecture with the GELU activation function. However, according to the authors of ALBERT, this model has more highlights than BERT comprising factorized embedding parameterization, cross-layer parameter sharing, and inter-sentence coherence loss. ALBERT achieves significantly better performances than BERT on various benchmark MRC corpora such as SQuAD and RACE \cite{lan2019albert}.

\end{itemize}


\subsection{Evaluation metrics}
Evaluations of MRC models on English and Chinese corpora \cite{Raj:16,Cui:18} used two evaluation metrics comprising exact match (EM) score and F1-score (macro-average). We also use these metrics to evaluate reading comprehension performances of humans and machine models on our corpus. In our Vietnamese MRC evaluations, punctuations and white spaces are ignored to allow normalization. The two evaluation metrics are described as follows.
\begin{itemize}
    \item {\bf EM} is the proportion of predicted answers that match exactly with the gold standard answers.
    \item {\bf F1-score} measures the overlap between the predicted answer and the gold standard answer. First, we generated the predicted answer and gold answer as sets of tokens, and then computed their F1-score. Finally, the evaluation system selected the maximum F1-score from all of the gold standard answers for each question, and then averaged them over all of the questions.
\end{itemize} 

\subsection{Experimental settings}
To adapt the DrQA \cite{Che:2017} and QANet \cite{Yu:18} systems to suit our corpus, we fine-tune parameters of the systems. In particular, we only use word features for all of the experiments with DrQA and QANet models. We use the pyvi tool\footnote[6]{We use the pyvi library https://pypi.org/project/pyvi/ for word segmentation.} for word segmentation. We employ 300-dimensional Pho2Vec \cite{tuan-nguyen-etal-2020-pilot} word embeddings as our pre-trained Vietnamese word embeddings for DrQA and QANet. We choose Pho2Vec because it was trained on the corpus including nearly 19GB data\footnote{https://github.com/binhvq/news-corpus} generated by removing similar articles and duplication from a 50GB Vietnamese news set, which contains the same domain as our corpus. Besides, we set \textit{batch size = 32} and \textit{epochs = 40} for both the two models.

We use the base-cased multilingual BERT and set of parameters as follows: 12 layers, 768 hidden dimensions, and 12 attention heads (in the transformer) with 179M parameters and a vocabulary of about 120k vocabulary. We select the best hyperparameters by searching a combination of the batch size, learning rate and the number of fine-tuning epochs from the following ranges: learning rate: {$2e-5$ , $3e-5$, $5e-5$}; batch size: {4, 8, 16, 32}; number of epochs: {2, 3}. The best hyperparameters and models are selected based on the performance with the development set in Vietnamese.

For the ALBERT experiments, we use a pre-trained\footnote{https://github.com/ngoanpv/albert\_vi} model for embedding input data. The training data for the embedding model is extracted from the Vietnamese texts with a vocabulary size of 30,000, tokenized by SentencePiece \cite{richardson-2018-sentencepiece}. Input data is processed in two forms: original (ALBERT$_{Base\_Cased}$) and lower case (ALBERT$_{Base\_Uncased}$) to increase the learning ability of the model through a combination of assessment of form and semantics of each word in the corpus based on the multi-meanings of the Vietnamese language. We follow the ALBERT baseline, and tune several parameters on the development set, and obtain the best results with learning rate: $2e-5$, batch size: 32 and epochs: 5. 

For the BERT and ALBERT experiments, based on our corpus characteristics, we choose the maximum answer length to 200, the question length to 50, and the input sequence length to 512.

\subsection{Estimation of human performance}
\label{humper}

To answer the question {\bf Q2}, we measure the performance of humans with the development and test sets for our corpus. In particular, we hire three other workers to independently answer questions using the test and development sets, with four answers per question, as described in the phase for collecting additional answers (in Section \ref{additionalanswer}). In contrast to Rajpurkar et al. \shortcite{Raj:16}, we use a cross-validation methodology to measure the performance of humans in a similar manner to and similar to Cui et al. \shortcite{Cui:18}. In particular, we consider the first answer as the human prediction and treated the remainder of the answers as ground truths. We obtain four human prediction performance results by iteratively treating the first, second, third, and fourth answers as the human predictions. We calculate the maximum the performance over all of the ground truth answers for each question. Finally, we average the four human prediction results as the final human performance result with on our corpus. Estimated human performances are presented in Table \ref{tab:result}.



\subsection{Experimental results}
Table \ref{tab:result} compares the performance of the machine models and humans with the development and test sets for our corpus. Random guess and Sliding Window are the two methods with the lowest results, achieving less than 15\% of the F1-score. Simple models cannot be used to deal with our data set. The transfer learning-based models (BERT  and ALBERT) significantly outperform better than the neural network-based models (DrQA and QANet), but not as well as humans. The best model (ALBERT$_{Base\_Uncased}$) achieves an EM of 65.26\% and a F1-score of 84.89\% on the test set. In particular, the performances (EM and F1-score) of the best model are better than that of DrQA, with 10.43\% in EM and 10.80\% in F1-score, respectively. The best model also perform better compared with the QANet model, where differences of 8.55\% (in EM) and differences 5.10\% (in F1-score). Meanwhile, the different performance between humans and the best model (14.53\% and 10.90\% differences in EM and F1-score, respectively) with our corpus was significant, thereby indicating that models for ViNewsQA should be improved in future research.

\begin{table}[!h]
\centering
\caption{Model and human performances on the development and test sets of ViNewsQA. Besides, we also build a approach based on random guess to compare with machine systems.}
\label{tab:result}
\setlength\arrayrulewidth{1pt}
\begin{tabular}{clrrrr}
\hline
\multirow{2}{*}{\textbf{Type}}                    & \multicolumn{1}{c}{\multirow{2}{*}{\textbf{MRC systems}}} & \multicolumn{2}{c}{\textbf{EM (\%)}}                                  & \multicolumn{2}{c}{\textbf{F1-score(\%)}}                             \\ \cline{3-6} 
                                                  & \multicolumn{1}{c}{}                                      & \multicolumn{1}{c}{\textbf{Dev}} & \multicolumn{1}{c}{\textbf{Test}} & \multicolumn{1}{c}{\textbf{Dev}} & \multicolumn{1}{c}{\textbf{Test}} \\ \hline
\multicolumn{1}{l}{\bf Random}                      & Random Guess                                               & 0.20                              & 0.15                               & 9.56                              & 9.30                               \\ \hline
\multicolumn{1}{l}{\bf Rule-based}                  & Sliding Window                                             & 0.32                              & 0.15                               & 13.11                             & 13.38                              \\ \hline
\multirow{2}{*}{\textbf{Neural Network-based}}    & DrQA                                                       & 49.26                             & 45.83                              & 74.03                             & 74.09                              \\
                                                  & QANet                                                      & 57.80                             & 56.71                              & 78.39                             & 79.79                              \\ \hline
\multirow{3}{*}{\textbf{Transfer learning-based}} & BERT                                                       & 64.56                             & 63.81                              & 81.47                             & 83.19                              \\ 
                                                  & ALBERT                                                     & 64.68                             & 64.46                              & 83.43                             & 84.16                              \\ 
                                                  & ALBERT                                                     & 64.24                             & 65.26                              & \textbf{83.52}                    & \textbf{84.89}                     \\ \hline
\multicolumn{2}{c}{\textbf{Human performance}}                                                               & 75.19                             & 79.79                              & 92.77                             & 95.79                              \\ \hline
\end{tabular}
\end{table}

                      

\section{Result analysis and discussion}
To obtain more insights into the performance of the neural network-based and transfer learning-based models and humans on our corpus, we analyze their performances in terms of different linguistic aspects comprising the question length (see in Section \ref{eff_ques_len}), answer length (see in Section \ref{eff_ans_len}), article length (see in Section \ref{eff_art_len}), question type (see in Section \ref{eff_ques_type}), answer type (see in Section \ref{eff_ans_type}) and reasoning type (see in Section \ref{eff_rea_type}). These aspects are described in Section 4. In addition, we also aim to examine the impacts of the size of the training set as well as the vocabulary size on the machine models (see in Section \ref{trainingsize}). Finally, we perform qualitative analysis through typical examples (see Section \ref{typicalexamples}). In this study, we omit analyzing these results on the lexical-based approach (Sliding Window) because its performance is significantly lower than other models.


\subsection{Effects of question length}
\label{eff_ques_len}
Firstly, we examine how well the reading comprehension models handle questions with different lengths. In particular, we analyze the performance of the machine models and humans in terms of the F1-score metric. Table \ref{tab:f1questionlengthanalysis} presents the detailed analysis of performances with various questions lengths. In general, more accurate results are obtained for long (>16 words) than short and average-length (<15 words) questions on the best model. The difference in performance between humans and the best model decreases as the question length increases. A plausible explanation for this 
is that longer questions tend to contain more sufficient information in order to be able to extract the correct answer span, which is similar to what on SQuAD \cite{wadhwa2018comparative} and UIT-ViQuAD \cite{Kiet:20ViQuAD}.

\begin{table}[H]
\caption{Performance in terms of F1-score (\%) according to the question length with the development set for our corpus. $\Delta$ is the performance difference between the human and the best-performance machine model (ALBERT).}\label{tab:f1questionlengthanalysis}
\setlength\arrayrulewidth{1pt}
\begin{tabular}{lrrrrrrr}
\hline
\multicolumn{1}{c}{{\bf Question length}} & {\bf DrQA} & {\bf QANet} &{\bf BERT} &{\bf ALBERT} & {\bf Human} & $\Delta$ \\ \hline
4-5     &72.56   &73.76    &{\bf 86.36}   &85.14  &95.60   &+10.46       \\ 
6-7    &71.68   &76.96    &80.43   &{\bf 82.59}  &92.95   &+10.36        \\ 
8-9   &74.69   &79.92    &79.46   &{\bf 82.01}  &92.91   &+10.90        \\ 
10-11   &74.98   &77.83    &82.33   &{\bf 84.28}  &92.68   &+8.40        \\
12-13     &73.18   &77.74    &83.85   &{\bf 85.00} &93.34   &+8.34       \\ 
14-15    &73.97   &78.00    &78.93   &{\bf 81.08} &92.03   &+10.95        \\ 
16-17   &77.07   &82.63    &81.70   &{\bf 85.85} &93.02   &+7.17        \\ 
18-19   &74.17   &70.74    &85.27   &{\bf 85.67} &94.65   &+8.98        \\
\textgreater{}19    &72.23   &88.03    &85.01   &{\bf 89.83} &93.78   &+3.95      \\ \hline
\end{tabular}
\end{table}

\subsection{Effects of answer length}
\label{eff_ans_len}
In order to examine how well the reading comprehension models could predict the answers with different lengths, we analyze the performances of the machine models and humans in terms of the EM and F1-score. As can be seen from Table \ref{tab:f1answerlengthanalysis}, our analysis shows the performances obtained with different answer lengths. In general, more accurate results are achieved for the shorter answer than longer answers. In particular, the best model achieves the highest performance with short answers (1--2 words, which accounts for 12.85\% of the corpus) and the lowest performance with long answers (>18 words, which accounts for 14.42\% of the corpus). In particular, answers with a length of 3 to 18 words witness fluctuation in performance. Hence, longer answers may be more difficult in finding answers. However, a difficult question is caused by many other factors such as such as reasoning type, answer type or question type. We examine this hypothesis in the following analysis.

\begin{table}[H]
\caption{Performance in terms of F1-score (\%) according to the answer length with the development set for our corpus. $\Delta$ is the performance difference between the human and the best-performance machine model (ALBERT).}\label{tab:f1answerlengthanalysis}
\setlength\arrayrulewidth{1pt}
\begin{tabular}{lrrrrrr}
\hline
\multicolumn{1}{c}{{\bf Answer length}} & {\bf DrQA} & {\bf QANet} & {\bf BERT} & {\bf ALBERT} & {\bf Human} & $\Delta$ \\ \hline
1-2     &73.55   &77.78    &83.95   &{\bf 85.09} &94.10   &+9.01        \\ 
3-4    &79.46   &76.69    &84.44   &82.73 &92.51   &+9.78      \\ 
5-6  &79.11   &75.07    &79.80   &83.52 &92.46   &+8.94        \\ 
7-8   &78.78   &78.77    &79.95   &84.08  &91.89   &+7.81       \\
9-10     &83.91   &79.43    &83.39   &83.83 &91.80   &+7.97        \\ 
11-12    &79.03   &79.89    &80.31   &83.16 &92.65   &+9.49        \\ 
13-14   &78.57   &80.86    &79.63   &84.32 &92.23   &+7.91        \\ 
15-16   &76.73  &82.96   &85.20  &84.91 &91.99   &+7.08        \\
17-18    &62.83  &80.46    &79.26   &84.55 &92.62   &+8.07       \\ 
19-20   &56.13   &77.70    &79.99  &{\bf 77.77} &91.80   &+14.03        \\ 
\textgreater{}20    &50.65   &79.36    &79.01   &{\bf 80.99} &91.87   &+10.88        \\\hline
\end{tabular}
\end{table}
\subsection{Effects of article length}
\label{eff_art_len}

In addition to determining the impacts of the question and answer lengths on the MRC model for the Vietnamese language, we analyze the performances of the MRC models and humans (in F1-score) with various article lengths. The detailed results are presented in and Table \ref{tab:f1articlelengthanalysis}. In general, more accurate results are obtained for shorter articles than longer articles. In particular, all the models have a tendency to achieve better performances with short articles (<301 words). Because longer articles have higher interference and take more time to process. Hence, the MRC system has difficulty finding the answer for longer articles, which is useful for future improvements.


\begin{table}[H]
\caption{Performance in terms of F1-score (\%) according to the article length with the development set for our corpus. $\Delta$ is the performance difference between the human and the best-performance machine model (ALBERT).}\label{tab:f1articlelengthanalysis}
\setlength\arrayrulewidth{1pt}
\begin{tabular}{lrrrrrr}
\hline
\multicolumn{1}{c}{{\bf Article length}} & {\bf DrQA} & {\bf QANet} & {\bf BERT} & {\bf ALBERT} & {\bf Human} & $\Delta$ \\ \hline
\textless{}101  &83.49   &89.85    &{\bf 98.00}   &82.44  &97.06   &+14.62        \\ 
101-200         &78.07   &81.45    &85.79   &{\bf 87.21} &94.93   &+7.72        \\
201-300         &74.25   &79.22    &82.63   &{\bf 84.57}  &93.20   &+8.63       \\
301-400         &71.83   &76.58    &79.26   &{\bf 82.37}  &92.87   &+10.50       \\
401-500         &72.17   &74.27    &78.55   &{\bf 79.36}  &90.49   &+11.13        \\
501-600         &74.01   &80.53    &81.33   &{\bf 82.94}  &92.37   &+9.43       \\
\textgreater{}600    &72.49   &76.76    &75.41   &{\bf 81.12} &91.94   &+10.82            \\ \hline
\end{tabular}
\end{table}

\subsection{Effects of question type}
\label{eff_ques_type}

We also analyze the performance of the model and humans in terms of the linguistic aspects based on the question type. Table \ref{tab:f1questiontypeanalysis} illustrates the detailed performances with different question types. In general, the "How," "Why," "What" and "Who" questions in our corpus are more difficult than others. In particular, ALBERT achieves the lowest performance on the question "Who". This is explained because "Who" has the lowest percentage of 3.16\% (see Table \ref{tab:questiontype1}) when compared to other types of questions in the corpus, and its answers with complex-structure noun phrases are difficult to locate the beginning and ending positions of answers (see the first example in Section \ref{peopleques}). The MRC system more readily extracts the correct answers for "Where", "When" and "Others" questions, and the best model achieves 88.16\%, 84.47\% and 89.14\%, respectively, because their answers were mostly short, from 1 to 5 words (see Table \ref{tab:answer-questiontype}). The differences in performance between humans and the best models are high for "Why", "How", "What", and "Who" questions (with differences in F1-scores over 9\%). These are the types of questions needed to improve their performance in the future.





\begin{table}[H]
\centering
\caption{Performance in terms of F1-score (\%) according to the  question type with the development set for our corpus. $\Delta$ is the performance difference between the human and the best-performance machine model (ALBERT).}
\label{tab:f1questiontypeanalysis}
\setlength\arrayrulewidth{1pt}
\begin{tabular}{lrrrrrr}
\hline
\multicolumn{1}{c}{{\bf Question type}} 
& \multicolumn{1}{c}{\textbf{DrQA}}
& \multicolumn{1}{c}{\textbf{QANet}}
& \multicolumn{1}{c}{\textbf{BERT}}
& \multicolumn{1}{c}{\textbf{ALBERT}}
& \multicolumn{1}{c}{\textbf{Human}} & \multicolumn{1}{c}{\textbf{$\Delta$}}\\ \hline
Who 	&66.64 &66.57 &{\bf 81.51}	&76.25 &95.00	&+18.75 \\

When	&76.07 &77.92 &82.10	&{\bf 84.47} &91.61	&+7.14 \\
Where	&72.96 &74.08 &83.12	&{\bf 88.16} &95.53	&+7.37 \\
Why 	&74.60 &76.45 &79.69	&{\bf 81.37}  &93.17	&+11.80 \\
What	&73.45 &78.87 &80.77	&{\bf 83.11}  &93.05	&+9.94 \\
How	    &70.82 &76.99 &78.10	&{\bf 83.15}  &91.66	&+8.52\\
Others	&83.04 &84.95 &{\bf 91.18}	&89.25  &92.93	&+3.68 \\ \hline
\end{tabular}

\end{table}

\subsection{Effects of answer type}
\label{eff_ans_type}
Besides, we also aim to verify the impacts of answer types to the MRC models. Table \ref{tab:f1answertypeanalysis} show the EM and F1-score performances according to different types of answer. "Person" and "Others" are two difficult answer types, achieving the lowest performances including 77.14\% (in F1-score) and 73.88\% (in F1-score), respectively. The 
Person answer type has the same results as for the Who question because they are closely related (see the second example in Section \ref{peopleques}). Besides, the Others answer type is very long compared to other types (see in Table \ref{tab:answer-answertype}), which is why it has low performance.


\begin{table}[H]
\caption{Performance in terms of the F1-score (\%) according to the answer type with the development set for our corpus. $\Delta$ is the performance difference between the human and the best-performance machine model (ALBERT).}\label{tab:f1answertypeanalysis}
\setlength\arrayrulewidth{1pt}
\begin{tabular}{lrrrrrr}
\hline
\multicolumn{1}{c}{{\bf Answer type}} & {\bf DrQA} & {\bf QANet} & {\bf BERT} & {\bf ALBERT} & {\bf Human} & $\Delta$ \\ \hline
Time  &77.47   &75.15    &82.43  &{\bf 85.18}    &93.33    &+8.15      \\ 
Other Numeric         &83.95   &86.74    &{\bf 90.80}   &89.27  &92.71   &+3.44       \\   
Person         &73.49   &{\bf 82.13}   &77.51   &77.14  &97.85   &+20.71       \\
Location         &52.66   &63.62    &85.67   &{\bf 85.83}  &98.47   &+12.64       \\
Other Entity         &67.76   &70.08    &81.84   &{\bf 86.66}  &93.28   &+6.62       \\
Noun Phrase         &75.20   &76.37    &82.13   &{\bf 83.44}  &93.02   &+9.58       \\
Verb Phrase    &72.93   &80.15    &81.50   &{\bf 82.94}  &93.16   &+10.22       \\ 
Adjective Phrase         &78.08   &81.18    &80.05   &{\bf 85.60}  &92.76   &+7.16      \\
Preposition Phrase         &77.39   &67.89    &64.43   &{\bf 81.29}  &94.13   &+12.84       \\
Clause    &72.37   &78.96    &80.40   &{\bf 84.92}  &92.23   &+7.31       \\ 
Others    &64.73   &73.22    &68.06   &{\bf 73.88}  &90.47   &+16.59      \\ 
\hline
\end{tabular}
\end{table}

\subsection{Effects of reasoning type}
\label{eff_rea_type}

We examine how well the MRC models could handle answers with different reasoning types. Table \ref{tab:f1reasoningtypeanalysis} shows the performance with different reasoning types. In general, more accurate results are gained for non-inference questions (word matching and paraphrasing) than inference questions (single-sentence reasoning and multi-sentence reasoning). This analysis provides clear insights that more complex inference forms make it difficult for MRC systems.

The differences in performance (F1-score) between humans and the best model are significant with over 11\% for inference questions, and with 3.3 to  8.4\% for simple questions.







\begin{table}[H]
\centering
\caption{Performance in  terms of F1-score (\%) according to the reasoning type with the development set for our corpus.}\label{tab:f1reasoningtypeanalysis}
\setlength\arrayrulewidth{1pt}
\resizebox{\columnwidth}{!}{\begin{tabular}{lrrrrrrr}
\hline
\multicolumn{1}{c}{{\bf Reasoning type}}
& \multicolumn{1}{c}{\textbf{DrQA}}
& \multicolumn{1}{c}{\textbf{QANet}}
& \multicolumn{1}{c}{\textbf{BERT}}
& \multicolumn{1}{c}{\textbf{ALBERT}}
& \multicolumn{1}{c}{\textbf{Human}} & \multicolumn{1}{c}{\textbf{$\Delta$}}\\ \hline
Word matching   &85.38 &88.02 &{\bf 92.75}	&91.28  &94.60	&+3.32 \\ 
Papaphrasing    &75.62 &80.73 &83.10	&{\bf 84.96}  &93.38	&+8.42 \\ 

Single-sentence reasoning       &71.50 &74.94 &75.38	&{\bf 81.08} &92.49	&+11.41 \\ 

Multi-sentence reasoning       &62.38 &67.83 &72.28	&{\bf 75.42} &91.18	&+15.76 \\ 

Ambiguous/Insufficient       &54.83 &64.49 &55.49	&{\bf 83.61} &86.51	&+2.90 \\

\hline
\end{tabular}}

\end{table}

\subsection{Effects of training set size}
\label{trainingsize}

\begin{figure}[H]
    \begin{minipage}{0.4\textwidth}
        \centering
        \captionof{table}{Vocabulary size of different amounts of training data.}\label{tab:vob}
        \setlength\arrayrulewidth{1pt}
        \begin{tabular}{lc}\hline
          {\bf \#Questions} & {\bf \#Vocabulary size} \\ \hline
          \multicolumn{1}{r}{1,992} &  \multicolumn{1}{r}{9,816}           \\ 
          \multicolumn{1}{r}{3,990}                & \multicolumn{1}{r}{13,935}          \\ 
\multicolumn{1}{r}{5,990}                & \multicolumn{1}{r}{17,076}          \\ 
\multicolumn{1}{r}{7,990}                & \multicolumn{1}{r}{19,763}          \\ 
\multicolumn{1}{r}{9,990}                & \multicolumn{1}{r}{22,079}          \\ 
\multicolumn{1}{r}{11,980}               & \multicolumn{1}{r}{24,137}          \\ 
\multicolumn{1}{r}{13,985}               & \multicolumn{1}{r}{26,062}          \\ 
\multicolumn{1}{r}{15,985}               & \multicolumn{1}{r}{27,880}          \\ 
\multicolumn{1}{r}{17,583}               & \multicolumn{1}{r}{29,142}          \\ \hline
          \end{tabular}
        \end{minipage}
        \hfill
    \begin{minipage}{0.5\textwidth}
        \centering
        \begin{tikzpicture}
        \begin{axis}[
            height=6.35cm,
            legend entries={ALBERT, BERT, QANet, DrQA},
            legend pos=south east,
            legend style={font=\tiny},
            legend cell align={left},
            xlabel near ticks,
            ylabel near ticks,
            xlabel style={font=\footnotesize},
            ylabel style={font=\footnotesize},
            xlabel=Number of questions,
            ylabel=F1-score (\%),
            xtick={1992,3990,5990,7990,9990,11989,13985,15985,17583},
            ymin=0,
            ymax=100,
            xticklabel style={
                rotate=45,
                font=\tiny,
                scaled x ticks = false,
            },
            yticklabel style={
                font=\tiny,
            },
            table/col sep=comma,
        ]
            \addplot table [x=t, y=albert_f1] {results.csv};
            \addplot table [x=t, y=bert_f1]  {results.csv};
            \addplot table [x=t, y=qanet_f1]  {results.csv};
            \addplot table [x=t, y=drqa_f1]  {results.csv};
        \end{axis}
    \end{tikzpicture}
    \end{minipage}
    \caption{Analysis and visualization of the model's result with different sizes of training data.}
    \label{fig:gradeanalysis}
\end{figure}


Our training data comprising 17,583 questions is much lower than the amount of the data used for English and Chinese MRC systems. To verify whether the small amount of training data affect the poor performance of the MRC systems different sizes training data including 1992, 3990, 5990, 7990, 9990, 11980, 13985, 15985, and 17583 questions. Table \ref{tab:vob} shows that increasing data size means increasing vocabulary size. Figure \ref{fig:gradeanalysis} presents the performance (F1-score) on the test set of the ViNewsQA corpus. In general, the performance of the systems is improved as the size of the training set increases from 1,992 to 17,583 questions. However, the F1-score increases  steadily when the number of questions increased from 7,999 to 17,583 with DrQA and QANet. These observations indicate that the transfer learning models (BERT and ALBERT) are more effective with a small amount of training data compared with the other two models. However, we find that their performances still increase with enhancing the size of the training set. Consequently, increasing the training data size can improve the performance of the MRC models and depend on the model type.

\subsection{Error analysis for typical examples}
\label{typicalexamples}

To obtain a better sense of what errors the reading comprehension systems are making on the ViNewsQA corpus, this work revolve two following questions: Q3: Why are the questions related to people not good at predicting models? and Q4: Which questions make neural network-based (DrQA and QANet) and transfer learning-based (BERT and ALBERT) systems all predict wrong?

\subsubsection{Questions related to people}
\label{peopleques}
In order to answer the question Q3, we observe on many examples and select two typical samples for analysis, described as follows.

\begin{itemize}

\item[] {\bf Context}: Ngày 6/12, kíp mổ gồm 15 bác sĩ và 15 y tá Bệnh viện SS ở Varanasi, bang Uttar Pradesh, đã tiến hành ca phẫu thuật miễn phí cho cặp song sinh dính liền. (\textit{On December 6, a team of 15 doctors and 15 nurses at SS Hospital in Varanasi, Uttar Pradesh state, performed the conjoined twins surgery for free.})
\item[] {\bf Question}: Ai phẫu thuật miễn phí? (\textit{Who does the surgery for free?})
\item[] {\bf Correct answer}: kíp mổ gồm 15 bác sĩ và 15 y tá Bệnh viện SS ở Varanasi, bang Uttar Pradesh (\textit{a team of 15 doctors and 15 nurses at SS Hospital in Varanasi, Uttar Pradesh state})
\item[] {\bf DrQA prediction}: 15 bác sĩ (\textit{15 doctors}) 
\item[] {\bf QANet prediction}: song sinh dính liền (\textit{conjoined twins})
\item[] {\bf BERT prediction}: kíp mổ gồm 15 bác sĩ và 15 y tá Bệnh viện SS ở Varanasi, bang Uttar Pradesh, đã tiến hành ca phẫu thuật miễn phí cho cặp song sinh dính liền (\textit{a team of 15 doctors and 15 nurses at SS Hospital in Varanasi, Uttar Pradesh state, performed the conjoined twins surgery for free})
\item[] {\bf ALBERT prediction}: cặp song sinh dính liền (\textit{the conjoined twins})

\end{itemize}

"Who" questions having complex-structured answers such as noun phrases are difficult to be predicted by the MRC systems. Nguyen et al. \cite{Kiet:20ViQuAD} also found that noun-phrase answers are not easy to extract. Therefore, predicted answers are incorrect or deemed.

\begin{itemize}

\item[] {\bf Context}: Thai phụ mang thai lần hai 37 tuần, em bé ngôi thuận có dấu hiệu chuyển dạ, được đưa vào Bệnh viện Phụ sản Hải Phòng đêm 29/5. Phó giáo sư Vũ Văn Tâm, Giám đốc Bệnh viện Phụ sản Hải Phòng trực tiếp mổ sinh, bé trai chào đời nặng 2,8 kg với dây rốn dài 50 cm thắt nút đơn như bím tóc tết. Bác sĩ Tâm cho biết thai nhi có nút thắt dây rốn hiện vẫn là khó khăn cho chẩn đoán trước sinh, kể cả với những chuyên gia đầu ngành trên thế giới.(\textit{Pregnant women with a second pregnancy of 37 weeks, her dominant baby showing signs of labor, was admitted to Hai Phong Obstetrics Hospital on the night of May 29. Associate Professor Vu Van Tam, Director of Hai Phong Obstetrics Hospital directly gave birth, the baby boy was born weighing 2.8 kg with a 50 cm umbilical cord tied in a single knot like a braid. Dr Tam said that the fetus with umbilical cord knot is still difficult for prenatal diagnosis, even with leading experts in the world.})
\item[] {\bf Question}: Ai là người thực hiện ca phẫu thuật? (\textit{Who did perform the surgery?})
\item[] {\bf Correct answer}: Phó giáo sư Vũ Văn Tâm (\textit{Associate Professor Vu Van Tam})
\item[] {\bf DrQA prediction}: những chuyên gia đầu ngành (\textit{leading experts}) 
\item[] {\bf QANet prediction}: Phó giáo sư Vũ Văn Tâm (\textit{Associate Professor Vu Van Tam})
\item[] {\bf BERT prediction}: Thai phụ mang thai (\textit{Pregnant women})
\item[] {\bf ALBERT prediction}: Thai phụ mang thai lần hai 37 tuần, em bé ngôi thuận (\textit{Pregnant women with a second pregnancy of 37 weeks, her dominant baby})

\end{itemize}

The correct answer is Associate Professor Vu Van Tam (Associate Professor Tam Van Vu) including a title (Associate Professor) and a human name (Vu Van Tam). Following this phrase in the context is a phrase that adds information about Associate Professor Vu, which may interfere with the predicted result. According to the analysis in Section \ref{eff_ans_type}, the QANet model better predicts person-entity-based answers.

\subsubsection{Predicted answers all models wrong} To answer the question Q4, we select and analyze the following context-question-answer triples, where neural network-based (DrQA and QANet) and transfer learning-based (BERT and ALBERT) systems all predict wrong.
\\

\textbf{Ambiguous answers: }Multiple spans in the context can be selected as correct answers for a question, so they create ambiguity in the process of predicting the correct answer by the MRC systems. An example is provided as follows.

\begin{itemize}

\item[] {\bf Context}: Theo báo cáo nhanh của Bộ Y tế, ngày 6/2 tức mùng 2 Tết nguyên đán Kỷ Hợi, Bệnh viện Bạch Mai (Hà Nội) ghi nhận 2 người viêm phổi nặng nghi nhiễm cúm gia cầm nguy hiểm. Hai trường hợp này đang được Viện Vệ sinh dịch tễ Trung ương điều tra dịch tễ và lấy mẫu xét nghiệm. Kết quả sẽ có trong vài ngày tới. Theo Cục Thú y, Bộ Nông nghiệp và Phát triển nông thôn, hiện cả nước không có ổ dịch cúm nào xảy ra trên gia cầm, nhưng không loại trừ khả năng cúm gia cầm xuất hiện trên người. Cũng theo Bộ Y tế, trong 6 ngày nghỉ Tết nguyên đán, không ghi nhận trường hợp nào mắc bệnh sởi và bệnh liên cầu lợn, chỉ một số ổ dịch sốt xuất huyết tại An Giang, Bà Rịa - Vũng Tàu và Bến Tre. Ngoài ra còn có một ổ dịch quai bị ở Bến Tre. (\textit{English translation: According to a quick report of the Ministry of Health, on February 6 the second day of the Ky Hoi Lunar New Year, Bach Mai Hospital (Hanoi) recorded 2 people with severe pneumonia suspected of being infected with dangerous avian influenza. These two cases are being investigated and tested by the Central Institute of Hygiene and Epidemiology. Results will be available in the next few days. According to the Department of Animal Health, Ministry of Agriculture and Rural Development, there is no flu outbreak in poultry nationwide, but it is not excluded that avian flu occurs in humans. According to the Ministry of Health, during the six days of the Tet holidays, no cases of measles and swine streptococcus were recorded, only a few outbreaks of dengue fever in An Giang, Ba Ria - Vung Tau, and Ben Tre. There is also a mumps outbreak in Ben Tre.})
\item[] {\bf Question}:  Ở Bến Tre có trường hợp nhiễm phải căn bệnh nào? (\textit{What disease is there in Ben Tre?})
\item[] {\bf Correct answer}: sốt xuất huyết. (\textit{dengue fever.}).
\item[] {\bf DrQA prediction}: bệnh sởi và bệnh liên cầu lợn. (\textit{measles and swine streptococcus.}).
\item[] {\bf QANet prediction}: viêm phổi nặng nghi nhiễm cúm gia cầm nguy hiểm.  (\textit{severe pneumonia suspected of being infected with dangerous avian influenza.}).
\item[] {\bf BERT prediction}: bệnh sởi và bệnh liên cầu lợn. (\textit{measles and swine streptococcus.}).
\item[] {\bf ALBERT prediction}: sởi và bệnh liên cầu lợn. (\textit{measles and swine streptococcus.}).

\end{itemize}

In this example, we find that two spans such as \textit{"sốt xuất quyết" (dengue fever)} and \textit{"quai bị"} (mumps) in the context are correct answers to the question \textit{"Ở Bến Tre có trường hợp nhiễm phải căn bệnh nào?"} (What disease is there in Ben Tre?). Hence, this question is ambiguous in finding the correct answer.
\\

\textbf{Incorrect boundary: }The answers predicted by the four machine models are roughly equivalent to the correct answers, even though they lack or excess some words compared with the correct answers. This error is also easily caused by humans. These predicted answers have no meaning in the EM evaluation; however, they are calculated into the F1-score evaluation. An example is given as follows.

\begin{itemize}
    \item[] \textbf{Context}:  Hút thuốc lào hay thuốc lá đều có hại cho sức khỏe của chính bản thân người hút và những người xung quanh. Thuốc lào có hàm lượng nicotin khoảng 9\%, cao hơn nhiều so với thuốc lá thông thường (khoảng 1- 3\%). (\textit{Tobacco or pipe tobacco smoking is both harmful to the health of the smokers and those around them. Pipe tobacco has a nicotine content of about 9\%, much higher than regular tobacco (about 1- 3\%).})
    \item[] \textbf{Question}:  Hút thuốc lá có hại như thế nào đối với con người? (\textit{How harmful is smoking for humans?}).
    \item[] \textbf{Correct answer}:  có hại cho sức khỏe của chính bản thân người hút và những người xung quanh. (\textit{harmful to the health of the smokers and those around them}).
    \item[] \textbf{DrQA prediction}:  bản thân người hút và những người xung quanh. (\textit{the health of the smokers and those around them}).
    \item[] \textbf{QANet prediction}:  đều có hại cho sức khỏe của chính bản thân người hút và những người xung quanh. (\textit{both harmful to the health of the smokers and those around them}).
    \item[] \textbf{BERT prediction}:  cho sức khỏe của chính bản thân người hút và những người xung quanh. (\textit{to the health of the smokers and those around them}).
    \item[] \textbf{ALBERT prediction}:  có hại cho sức khỏe. (\textit{harmful to the health}).

\end{itemize}

In this question, compared with the correct answer, the DrQA, QANet, BERT answer predictions lack or excess a word and several words. However, these answers (DrQA, QANet, and BERT predictions) are deemed, which could be accepted as the correct answers.
\\

\textbf{Incorrect inference: }According to the analysis in Sub-section 6.5, the questions inferred based on single-or-multiple-sentence information have lower results than the non-inference questions. We select an example for this error, as described follows.

\begin{itemize}
    \item[] \textbf{Context}:  Theo SCMP, các bác sĩ từng cảnh báo Wu Ying "hoàn toàn không phù hợp để mang thai" vì bệnh tim bẩm sinh cùng chứng tăng huyết áp phổi. Chồng Wu là Shen Jie cũng sẵn sàng từ bỏ đứa bé để cứu vợ, song Wu vẫn kiên quyết giữ con. "Nhiều người bảo tôi cứng đầu nhưng họ đều đã có con rồi", Wu trải lòng. "Mỗi lần nhìn con cái họ, tôi lại muốn sinh ra đứa con của chính mình. Tôi hiểu rất rõ các rủi ro nhưng sẵn sàng đánh cược". Wu từng sảy thai hai lần. Tháng 5/2017, Wu sinh con trai nặng một kg bằng phương pháp đẻ mổ. (\textit{According to SCMP, doctors warned Wu Ying "completely unsuitable for pregnancy" because of congenital heart disease and pulmonary hypertension. Wu's husband Shen Jie is also willing to give up the baby to save his wife, but Wu is determined to keep the child. "Many people told me to be stubborn, but they all have children," Wu said. "Every time I see their children, I want to give birth to my own child. I understand the risks very well but am willing to bet." Wu had miscarried twice. In May 2017, Wu gave birth to a one-kilogram son by cesarean section.}) 
    \item[] \textbf{Question}:  Vì sao Wu Ying vẫn quyết định giữ đứa bé? (\textit{Why did Wu Ying still decide to keep the baby?}).
    \item[] \textbf{Correct answer}:  muốn sinh ra đứa con của chính mình. (\textit{want to give birth to my own child.}).
    \item[] \textbf{DrQA prediction}:  Nhiều người bảo tôi cứng đầu nhưng họ đều đã có con rồi. (\textit{Many people told me to be stubborn, but they both had children.}).
    \item[] \textbf{QANet prediction}: bệnh tim bẩm sinh cùng chứng tăng huyết áp phổi. Chồng Wu là Shen Jie cũng sẵn sàng từ bỏ đứa bé để cứu vợ. (\textit{congenital heart disease and pulmonary hypertension. Wu's husband Shen Jie is also willing to give up the baby to save his wife.}).
    \item[] \textbf{BERT prediction}:  Nhiều người bảo tôi cứng đầu nhưng họ đều đã có con rồi. (\textit{Many people told me to be stubborn, but they all have children.}).
    \item[] \textbf{ALBERT prediction}:  vì bệnh tim bẩm sinh cùng chứng tăng huyết áp phổi. (\textit{because of congenital heart disease and pulmonary hypertension.})

\end{itemize}

To answer the question above, the human or MRC systems must understand and connect the contents of the first four sentences in the context to find the correct answer. Therefore, the predicted answers are chosen from a span of the first three sentences instead of a segment from the content of the $4^{th}$ sentence.
\\

\textbf{Lack of world knowledge: } Several questions require knowledge to determine the answers. For example, these questions may use equivalent words or phrases which can be terms or concepts of a specific field. We select an example for this error, as described follows.

\begin{itemize}
    \item[] {\bf Context}: Theo tiến sĩ Đào Văn Long, nguyên Trưởng khoa Tiêu hóa Bệnh viện Bạch Mai, người bị đau dạ dày thường do vi khuẩn HP, stress, lạm dụng chất kích thích và thói quen ăn uống không hợp lý. Chế độ ăn của người đau dạ dày cần giảm tác dụng của axít lên niêm mạc dạ dày, hạn chế hoặc bỏ những kích thích có hại để dạ dày nghỉ ngơi và các tổn thương mau lành. Do đó, người đau dạ dày cần ăn đúng giờ, hạn chế chất kích thích và chọn gia vị phù hợp. "Người đau dạ dày cần ăn uống nghiêm ngặt và cầu kỳ", tiến sĩ Long khuyên. (\textit{According to Dr. Dao Van Long, the former head of the Department of Gastroenterology at Bach Mai Hospital, people with stomach pain are often caused by HP bacteria, stress, substance abuse and inappropriate eating habits. The diet of the stomach ache should reduce the effect of the acid on the gastric mucosa, limit or eliminate harmful stimuli to rest the stomach and heal the damage. Therefore, people with stomach ache need to eat on time, limit stimulants and choose the right spices. "People with stomach ache need strict and sophisticated diet", Dr. Long advised.})
    \item[] {\bf Question}:  Người bị đau dạ dày cần kiêng những gì? (\textit{What should people with stomachache abstain from?})
    \item[] {\bf Correct answer}:  giảm tác dụng của axít lên niêm mạc dạ dày, hạn chế hoặc bỏ những kích thích có hại để dạ dày nghỉ ngơi và các tổn thương mau lành (\textit{reduce the effect of the acid on the gastric mucosa, limit or eliminate harmful stimuli to rest the stomach and heal the damage}).
    \item[] {\bf DrQA prediction}:  ăn uống nghiêm ngặt và cầu kỳ (\textit{strict and sophisticated diet}).
    \item[] {\bf QANet prediction}:  nghiêm ngặt và cầu kỳ (\textit{strict and sophisticated diet}).
    \item[] {\bf BERT prediction}:  ăn uống nghiêm ngặt và cầu kỳ (\textit{strict and sophisticated diet}).
    \item[] {\bf ALBERT prediction}:  ăn đúng giờ, hạn chế chất kích thích và chọn gia vị phù hợp (\textit{eat on time, limit stimulants and choose the right spices}).

\end{itemize}

In the context, there are three advices from a doctor that should be improved if you encounter stomachache. Therefore, this question is ambiguous to find the answer. If the systems has knowledge of the verb "kiêng" ("kiêng" means to avoid eating, not consuming certain foods or doing certain things, because it is harmful or considered harmful to health.), it is easy to find the correct answer from the answer candidates.
\section{Conclusion and future work}
This paper introduced ViNewsQA, a span-extraction corpus for evaluating intelligent reading comprehension systems and question-answering in a low-resource language like Vietnamese. Over 22,000 question-answer pairs were generated by humans based on a set of 4,416 online health news articles in our corpus. Our corpus contains diverse answer types, and a significant proportion of questions (42.25\% of ViNewsQA) required complex reasoning ability to solve. The corpus is challenging because our evaluation results showed that the difference in performance between humans and the best model was significant (an EM difference of 14.53\% and an F1-score difference of 10.90\%). Analyses of the experimental results showed that better performances were obtained for long questions with more information than short questions, whereas shorter answers and articles tend to yield better performances. Additionally, we realized that our corpus has difficult question types (What, How, Why, and Who) and complex reasoning based on a sentence  or connections between multiple sentences. Finally, we explored the qualitative analysis errors consisting of ambiguous answers, incorrect boundaries, incorrect inference and lack of world knowledge. 

By its size and complexity, ViNewsQA makes a significant extension to the existing machine reading comprehension corpora. For example, our corpus can be used for cross-lingual studies based on experiments with other similar corpora, such as SQuAD, NewsQA, and CMRC. We hope that our corpus will stir more research in machine reading comprehension and guide the development of artificial-intelligence applications. In particular, we conduct further investigations to solve the difficult questions that require comprehensive reasoning based on multiple sentences in the article. Moreover, we would like to operate a Vietnamese MRC challenging shared task for researchers to conduct experiments to explore better models with our corpus. Finally, we aim to build a modern QA system based on DrQA \cite{Che:2017} which is necessary to serve the searching demands of nearly 100M Vietnamese people.

\begin{acks}
We would like to thank the editors and anonymous reviewers for their helpful feedback. 

\end{acks}


\printbibliography

\end{document}